\documentclass[runningheads]{article}
\usepackage[left=3.0cm,right=3.0cm,top=2.5cm,bottom=2.5cm]{geometry} 
\usepackage{makeidx}
\usepackage{graphicx}
\usepackage{amsmath,amssymb,bbold} 
\usepackage{color}
\usepackage{listings}

\usepackage{bm}
\usepackage{algorithm}
\usepackage{algorithmicx}
\usepackage{rotating}
\usepackage{multirow}
\usepackage{stackrel}
\usepackage{booktabs} 
\usepackage{pdflscape}
\usepackage{amsthm} 
\usepackage{overpic}
\usepackage{algpseudocode}
\usepackage{tabularx}
\usepackage{mathtools}
\usepackage[colorlinks=true,linkcolor=blue,citecolor=blue]{hyperref}

\newcommand{\Div}[1]{\operatorname{Div} \left ( #1 \right )}
\newcommand {\mathsym}[1]{{}}
\newcommand {\unicode}[1]{{}}
\newcommand{\imgsize}{0.16\linewidth}

\renewcommand{\div}[1]{\operatorname{div} \left ( #1 \right )}

\usepackage[uppercase,sc,center]{titlesec}

\newcommand{\ie}{i.e.}
\newcommand{\eg}{e.g.}

\renewcommand{\imgsize}{0.24\linewidth}
\newcommand{\overlayZoom}[3]{ 
  \begin{tikzpicture}
    \begin{scope}[spy using outlines={rectangle,magnification=2,size=1.5cm,color=red}] 
      \node[inner sep=0,outer sep=0,anchor=south west] (image) at (0,0) 
      {\includegraphics[width=\imgsize]{#1}};
      \spy on (#2,#3) in node (zoom) at (2.95,0.8);
    \end{scope}
  \end{tikzpicture}
  \hspace*{-3mm}
}

\usepackage{tikz}
\usetikzlibrary{spy,calc}

\tikzset{annot/.style={draw=black,fill=white,text=black}}

\newif\ifblackandwhitecycle
\gdef\patternnumber{0}

\pgfkeys{/tikz/.cd,
    zoombox paths/.style={
        draw=orange,
        very thick
    },
    black and white/.is choice,
    black and white/.default=static,
    black and white/static/.style={ 
        draw=white,   
        zoombox paths/.append style={
            draw=white,
            postaction={
                draw=black,
                loosely dashed
            }
        }
    },
    black and white/static/.code={
        \gdef\patternnumber{1}
    },
    black and white/cycle/.code={
        \blackandwhitecycletrue
        \gdef\patternnumber{1}
    },
    black and white pattern/.is choice,
    black and white pattern/0/.style={},
    black and white pattern/1/.style={    
            draw=white,
            postaction={
                draw=black,
                dash pattern=on 2pt off 2pt
            }
    },
    black and white pattern/2/.style={    
            draw=white,
            postaction={
                draw=black,
                dash pattern=on 4pt off 4pt
            }
    },
    black and white pattern/3/.style={    
            draw=white,
            postaction={
                draw=black,
                dash pattern=on 4pt off 4pt on 1pt off 4pt
            }
    },
    black and white pattern/4/.style={    
            draw=white,
            postaction={
                draw=black,
                dash pattern=on 4pt off 2pt on 2 pt off 2pt on 2 pt off 2pt
            }
    },
    zoomboxarray inner gap/.initial=5pt,
    zoomboxarray columns/.initial=2,
    zoomboxarray rows/.initial=2,
    subfigurename/.initial={},
    figurename/.initial={zoombox},
    zoomboxarray/.style={
        execute at begin picture={
            \begin{scope}[
                spy using outlines={%
                    zoombox paths,
                    width=\imagewidth / \pgfkeysvalueof{/tikz/zoomboxarray columns} - (\pgfkeysvalueof{/tikz/zoomboxarray columns} - 1) / \pgfkeysvalueof{/tikz/zoomboxarray columns} * \pgfkeysvalueof{/tikz/zoomboxarray inner gap} -\pgflinewidth,
                    height=\imageheight / \pgfkeysvalueof{/tikz/zoomboxarray rows} - (\pgfkeysvalueof{/tikz/zoomboxarray rows} - 1) / \pgfkeysvalueof{/tikz/zoomboxarray rows} * \pgfkeysvalueof{/tikz/zoomboxarray inner gap}-\pgflinewidth,
                    magnification=3,
                    every spy on node/.style={
                        zoombox paths
                    },
                    every spy in node/.style={
                        zoombox paths
                    }
                }
            ]
        },
        execute at end picture={
            \end{scope}
     \gdef\patternnumber{0}
        },
        spymargin/.initial=0.5em,
        zoomboxes xshift/.initial=1,
        zoomboxes right/.code=\pgfkeys{/tikz/zoomboxes xshift=1},
        zoomboxes left/.code=\pgfkeys{/tikz/zoomboxes xshift=-1},
        zoomboxes yshift/.initial=0,
        zoomboxes above/.code={
            \pgfkeys{/tikz/zoomboxes yshift=1},
            \pgfkeys{/tikz/zoomboxes xshift=0}
        },
        zoomboxes below/.code={
            \pgfkeys{/tikz/zoomboxes yshift=-1},
            \pgfkeys{/tikz/zoomboxes xshift=0}
        },
        zoomboxes inside/.code={
            \pgfkeys{/tikz/zoomboxes yshift=-0.5},
            \pgfkeys{/tikz/zoomboxes xshift=0.5}
        },
        caption margin/.initial=4ex,
    },
    adjust caption spacing/.code={},
    image container/.style={
        inner sep=0pt,
        at=(image.north),
        anchor=north,
        adjust caption spacing
    },
    zoomboxes container/.style={
        inner sep=0pt,
        at=(image.north),
        anchor=north,
        name=zoomboxes container,
        xshift=\pgfkeysvalueof{/tikz/zoomboxes xshift}*(\imagewidth+\pgfkeysvalueof{/tikz/spymargin}),
        yshift=\pgfkeysvalueof{/tikz/zoomboxes yshift}*(\imageheight+\pgfkeysvalueof{/tikz/spymargin}+\pgfkeysvalueof{/tikz/caption margin}),
        adjust caption spacing
    },
    calculate dimensions/.code={
        \pgfpointdiff{\pgfpointanchor{image}{south west} }{\pgfpointanchor{image}{north east} }
        \pgfgetlastxy{\imagewidth}{\imageheight}
        \global\let\imagewidth=\imagewidth
        \global\let\imageheight=\imageheight
        \gdef\columncount{1}
        \gdef\rowcount{1}
        
    },
    image node/.style={
        inner sep=0pt,
        name=image,
        anchor=south west,
        append after command={
            [calculate dimensions]
            node [image container,subfigurename=\pgfkeysvalueof{/tikz/figurename}-image] {\phantomimage}
            node [zoomboxes container,subfigurename=\pgfkeysvalueof{/tikz/figurename}-zoom] {\phantomimage}
        }
    },
    color code/.style={
        zoombox paths/.append style={draw=#1}
    },
    connect zoomboxes/.style={
    spy connection path={\draw[draw=none,zoombox paths] (tikzspyonnode) -- (tikzspyinnode);}
    },
    help grid code/.code={
        \begin{scope}[
                x={(image.south east)},
                y={(image.north west)},
                font=\footnotesize,
                help lines,
                overlay
            ]
            \foreach \x in {0,1,...,9} { 
                \draw(\x/10,0) -- (\x/10,1);
                \node [anchor=north] at (\x/10,0) {0.\x};
            }
            \foreach \y in {0,1,...,9} {
                \draw(0,\y/10) -- (1,\y/10);                        \node [anchor=east] at (0,\y/10) {0.\y};
            }
        \end{scope}    
    },
    help grid/.style={
        append after command={
            [help grid code]
        }
    },
}

\newcommand\phantomimage{%
    \phantom{%
        \rule{\imagewidth}{\imageheight}%
    }%
}
\newcommand\zoombox[2][]{
    \begin{scope}[zoombox paths]
        \pgfmathsetmacro\xpos{
            (\columncount-1)*(\imagewidth / \pgfkeysvalueof{/tikz/zoomboxarray columns} + \pgfkeysvalueof{/tikz/zoomboxarray inner gap} / \pgfkeysvalueof{/tikz/zoomboxarray columns} ) + \pgflinewidth
        }
        \pgfmathsetmacro\ypos{
            (\rowcount-1)*( \imageheight / \pgfkeysvalueof{/tikz/zoomboxarray rows} + \pgfkeysvalueof{/tikz/zoomboxarray inner gap} / \pgfkeysvalueof{/tikz/zoomboxarray rows} ) + 0.5*\pgflinewidth
        }
        \edef\dospy{\noexpand\spy [
            #1,
            zoombox paths/.append style={
                black and white pattern=\patternnumber
            },
            every spy on node/.append style={#1},
            x=\imagewidth,
            y=\imageheight
        ] on (#2) in node [anchor=north west] at ($(zoomboxes container.north west)+(\xpos pt,-\ypos pt)$);}
        \dospy
        \pgfmathtruncatemacro\pgfmathresult{ifthenelse(\columncount==\pgfkeysvalueof{/tikz/zoomboxarray columns},\rowcount+1,\rowcount)}
        \global\let\rowcount=\pgfmathresult
        \pgfmathtruncatemacro\pgfmathresult{ifthenelse(\columncount==\pgfkeysvalueof{/tikz/zoomboxarray columns},1,\columncount+1)}
        \global\let\columncount=\pgfmathresult
        \ifblackandwhitecycle
            \pgfmathtruncatemacro{\newpatternnumber}{\patternnumber+1}
            \global\edef\patternnumber{\newpatternnumber}
        \fi
    \end{scope}
}

\numberwithin{equation}{section}
\numberwithin{figure}{section}

\theoremstyle{plain}
\newtheorem{theorem}{Theorem}[section]
\newtheorem{lemma}[theorem]{Lemma}
\newtheorem{proposition}[theorem]{Proposition}

\theoremstyle{definition}
\newtheorem{definition}{Definition}[section]

\newcommand{\bitem}{\begin{itemize}}
\newcommand{\eitem}{\end{itemize}}
\newcommand{\mc}[1]{\mathcal{#1}}

\newcommand{\N}{\mathbb{N}}
\newcommand{\R}{\mathbb{R}}

\newcommand{\bpm}{\begin{pmatrix}}
\newcommand{\epm}{\end{pmatrix}}
\newcommand{\bsm}{\left(\begin{smallmatrix}}
\newcommand{\esm}{\end{smallmatrix}\right)}
\newcommand{\T}{\top}

\newcommand{\la}{\langle}
\newcommand{\ra}{\rangle}

\newcommand{\mrm}[1]{\mathrm{#1}}

\newcommand{\vphi}{\varphi}

\newcommand{\eins}{\mathbb{1}}

\DeclareMathOperator{\Diag}{Diag}
\DeclareMathOperator{\diag}{diag}

\DeclareMathOperator{\vecmat}{vec}

\DeclareMathOperator{\TV}{TV}

\begin{document}
\pagestyle{headings}


\title{\textsc{A Geometric Approach \\ to Color Image Regularization}}

\author{\textsc{F. {\AA}str{\"o}m and C. Schn{\"o}rr}}

\date{}
\maketitle

\newcommand\blfootnote[1]{%
  \begingroup
  \renewcommand\thefootnote{}\footnote{#1}%
  \addtocounter{footnote}{-1}%
  \endgroup
}
\blfootnote{
 \hspace{-4.2mm}  \emph{Date: \today}


 (F. {\AA}str{\"o}m) Heidelberg Collaboratory for Image Processing, Heidelberg University, Germany 

  (C. Schn{\"o}rr) Image and Pattern Analysis Group, Heidelberg University, Germany 

  Support by the German Research Foundation (DFG) is gratefully acknowledged, grant GRK 1653.}

\begin{abstract}
  We present a new vectorial total variation method that addresses the problem of color consistent image filtering. Our approach is inspired from the double-opponent cell representation in the human visual cortex. Existing methods of vectorial total variation regularizers have insufficient (or no) coupling between the color channels and thus may introduce color artifacts. We address this problem by introducing a novel coupling between the color channels related to a pullback-metric from the opponent space to the data (RGB color) space. Our energy is a non-convex, non-smooth higher-order vectorial total variation approach and promotes color consistent image filtering via a coupling term. For a convex variant, we show well-posedness and existence of a solution in the space of vectorial bounded variation. For the higher-order scheme we employ a half-quadratic strategy, which model the non-convex energy terms as the infimum of a sequence of quadratic functions. In experiments, we elaborate on traditional image restoration applications of inpainting, deblurring and denoising. Regarding the latter, we demonstrate state of the art restoration quality with respect to structure coherence and color consistency. 
\end{abstract}

\tableofcontents

\section{Introduction}

\subsection{Motivation}
Image filtering is a fundamental operation in image processing applications. Typically image filtering refers to all type of algorithms that modify image pixels in a linear or non-linear manner. Common applictions are image denoising (or smoothing) \cite{Iijima,Koenderink1984,Perona90scale-spaceand,710815,diva2:543914,doi:10.1137/090769521,1467533}, active contours \cite{Kass88snakes:active}, image deblurring \cite{Oliveira20091683,661187}, inpainting \cite{935036,Chan02euler'selastica}  and optical flow \cite{Horn81determiningoptical}. These applications have in common that they can be formulated as variational problems and are thus inherently related. 

In a \emph{discrete setting}, the solution of such functionals (or energies), can be formulated as maximum a-posteriori (MAP) problems based on markov random fields (MRF), we refer to \cite{Wang2013b} and \cite{Kappes2015} for such approaches. However, the size of the required label space makes the optimization problem intractable as there is one label for each possible state. Due to this drawback, one computes approximate solutions, \eg, via $\alpha$-expansion or other relaxation techniques of the label space. The advantage of structured energy minimization, such as the MRFs formulation, is that complex neighborhoods, non-smooth and non-convex penalty functions are easily modelled.

On the other hand, \emph{continuous models} do not suffer from large label spaces, see for example the recently introduced assignment filter \cite{assignmentFilter}. However, the corresponding optimization problem needs to explicitly cope with non-smoothness and non-convexity. Convex optimization techniques are well established methods that efficiently find optimal solutions of convex functions.  During the past years, the imaging community has seen a surge of non-convex and often non-smooth energies, often demonstrating improved results over convex counterparts. The optimization of non-convex functions is particularly challenging since straightforward approaches often leads to locally optimal solution only. 

\begin{figure*}[t]
  \centering
  \hfill
  \includegraphics[width=0.25\linewidth]{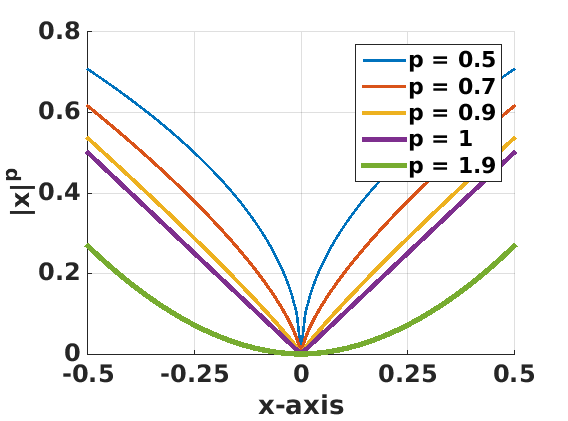}
  \hfill\hfill
  \includegraphics[width=0.24\linewidth]{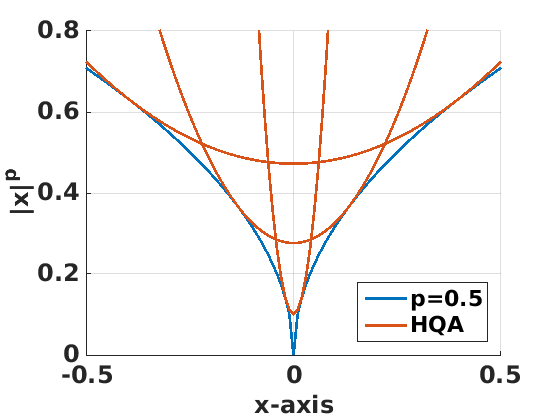}
  \hfill \phantom{I} \\
  \begin{tabularx}{1\linewidth}{XX}
    (\textbf{a}) Example of different values of the $p$-norm for $p\in (0,2)$. 
    & 
    (\textbf{b}) The non-convex energy realized as the infimum of quadratic functions (here $p=0.5$).
  \end{tabularx} \\[3mm]
  \renewcommand{\imgsize}{0.12\linewidth}
  \hfill
  \includegraphics[width=\imgsize]{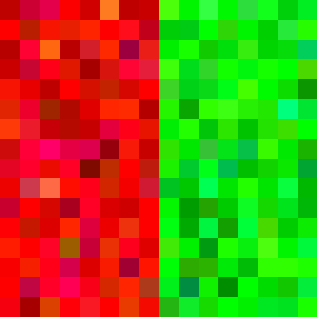}  
  \hspace{5mm}
  \includegraphics[width=\imgsize]{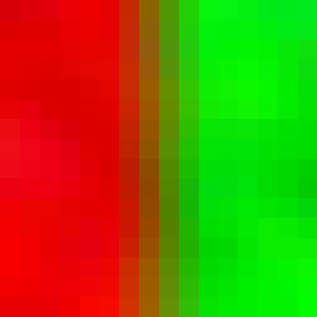}
  \hfill\hfill
  \includegraphics[width=\imgsize]{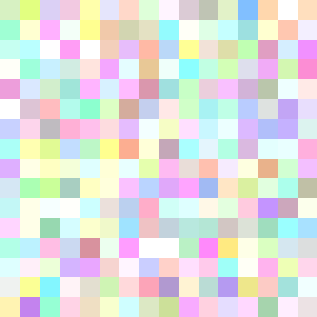} 
  \hspace{5mm}
  \includegraphics[width=\imgsize]{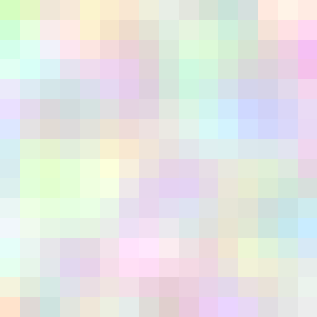} 
  \hfill\phantom{I}\\[1mm]
  \begin{tabularx}{1\linewidth}{XX}
    (\textbf{c}) Introduction of artificial colors are shown by the smooth transition from red to green in the right figure. These types of artifacts often arise in image smoothing due to insufficient color channel coupling and the denoising scheme's lack of adaptivity to the image structure.
    & 
    (\textbf{d}) Example of color shimmering artifacts (right) which often appear in homogeneous image regions. Color shimmering can often be suppressed with stronger smoothing, however, frequently at the cost of oversmoothing image features (corners/lines etc). 
  \end{tabularx}
  \caption{Figure (a) shows several instances of the non-convex $\ell_p$-norm for different $p$ values and (b) shows example HQA approximation for $p=0.5$. Figures (c) and (d) illustrates common color image denoising artifacts. }
  \label{fig:img_overview}
\end{figure*}

Relaxation of the non-smooth problems often include modification of the objective function and approximating non-convex penalty terms via auxiliary variables. Cohen proposed fitting of auxiliary variables \cite{Cohen:1996:AVT:226128.226132}. However, this approach relies on conjugate functions and if no closed form-solutions are available the relaxation method is inefficient. Another popular approach in image processing is the half-quadratic algorithm (HQA) introduced by Geman and Reynolds \cite{392335}. The HQA approximates a non-convex function as the infimum of quadratic functions as illustrated in Figure \ref{fig:img_overview} (a), (b). One may also consider lagged fix-point formulations \cite{doi:10.1137/S0036142997327075}. In this case regularity is imposed via mollification that yields a differentiable energy. Subsequently one needs to prove that there exists a convergent fixed-point algorithm. 

This work builds on the convex total variation (TV) presented in the seminal work of Rudin, Osher and Fatemi \cite{Rudin1992}. The success story of total variation (TV) began in 1992 when Rudin, Osher and Fatemi \cite{Rudin1992} introduced an extension of Rudin's PhD thesis \cite{RudinPhD}. In Rudin's work it was conjectured that the $\ell_1$ norm is more appropriate as a regularizer for image processing applications than, \eg, $\ell_2$ norm. The popularity of TV is mainly due to its discontinuity preserving properties, \ie, the norm is a strong prior for avoiding mode mixing and can be interpreted as a the solution of a MAP problem. The common  goal for noise reduction methods is to preserve characteristic image features, thus TV is a suitable prior as it is edge preserving. Features of interest vary depending on application area. However, in general one wishes to preserve structures defining dominant orientations and discontinuity points, such as edge and corners, since much of the visual information is contained in contour and differences of contrast \cite{Ratliff1971}. Extensions of the initial gray-scale TV prior for image enhancement to color images faces the problem to characterize notions of \emph{color}. The problem of consistent color image processing is largely unsolved and still no consensus on suitable characterization of a ``color edge'', or a ``color boundary'' for general imaging problems has been reached.

This work studies the problem of color image regularization. Extending the scalar TV to color images is a non-trivial problem. For example, if a color edge is insufficiently preserved in the smoothing process, artificial colors may emerge at the smooth transition between these colors as demonstrated by Figure \ref{fig:img_overview} (c). The same figure (d) illustrates the problem of color shimmering, \ie, insufficient smoothing of homogeneous regions. To address these problems we examine a color space representation, commonly used in computer vision applications and derive a novel color mixing term that penalizes inter-channel discontinuities. We investigate a special instance of color space representation known as the \emph{double-opponent color space}. The key aspect of our framework builds on the observation that the Jacobian carries vital information useful for color boundary detection. Utilizing this information, we design a TV-based regularizer that describes the color information in a subspace defined by the hue and saturation of the original image color space. Via a higher-order non-convex, non-smooth energy formulation we show improved discontinuity preserving properties over convex counter-parts with respect to color consistency and structural coherence.

Our approach is motivated based on results from color perception:
\begin{itemize}
\item The connection between experienced visual stimuli and current color space models of the visual cortex is naturally modeled using tools from differential geometry. Accordingly, we adopt a geometric viewpoint to explore the relation between color edges and the regularizer based on the color space geometries. The double-opponent color space is thought to relate neurophysiological properties of color experience to single-opponent and double-opponent cells in the human cortex, see \cite{6751225,Land83,Land86,Ebner:2007:CC:1535382} and references therein. There is recent evidence that a large concentration of double-opponent cells are located in the region V1, the primary part of the visual cortex \cite{Conway10}. Double-opponent cells are thought to be orientation-selective with respect to color discrimination and the detection of color boundaries, results made possible by modern functional magnetic resonance imaging (fMRI) techniques \cite{Conway10}. We will use this fact in our subsequent analysis to motivate the introduction of our model.
\end{itemize}

When formulating image denoising objective functions one often adopts different viewpoints. The following two major viewpoints motivate our work: namely color perception and color model. 
\begin{itemize}
\item \emph{Color perception}. As stated, we formulate the problem of color image denoising from principles of color perception. The discriminate power of color is one primary feature for object separation and detection. It is often referred to as a highly important features for the visual system and is closely related to the problem of accurate boundary detection \cite{Conway10}.  We present a model that preserves discontinuities in the color space motivated by a double-opponent transformation. By preserving color discontinuities we hypothesize that color borders trigger the activation of these double-opponent cells and thus yields the experience of crisp color borders in the image. 

\item \emph{Color Model}. We denote transportation of the visual (RGB) stimuli to the double-opponent cells in the visual cortex with a mapping. We postulate that, if there exists a spatial relation between two stimuli (\eg, a color difference), then this induces a response in the double-opponent cells in the form of orientation sensitivity. The motivation is that double-opponent cells act as color edge detectors, as shown by neurophysical experiments (again we refer to \cite{6751225,Land83,Land86,Ebner:2007:CC:1535382}). Thus, we conclude that there exist a color transition function (or gradient) in the double-opponent space. To obtain the mapping we observe that the stimuli in the opponent space, induced by a linear opponent transformation, in fact gives rise to a pullback metric on the RGB-space where the spatial interaction between the double-opponent cells are modeled by the gradient-operator. 

\end{itemize}
  
\subsection{Organization}
In Section \ref{sec:hist_TV} we sketch the framework of total variation and make the difference between our view-point to established literature in the field. Already here, we must emphasize that much research in color image processing is merely a multi-dimensional extension of the original total variation for gray-scale images. The differences between our approach and related VTV methods are also detailed in Section \ref{sec:hist_TV}. In Section \ref{sec:appearance} we review color space models often adopted in the image processing literature. In the same section we also introduce the double-opponent transformation. Section \ref{sec:metric_tensor} serves to derive the connection between the observation (RGB) space and the double-opponent representation, we also derive results on the encoded information in the metric decomposition and relate these facts to \emph{colorfulness}. The general variational problem is defined in Section \ref{sec:energy}. In the same section we formulate the corresponding HQA formulation and prove that the HQA is a particular instance of a majorize-minimize algorithm for the general problem. For the particular instance corresponding to the non-relaxed, first order VTV with a convex dataterm, we rigorously show convexity of the overall problem, that a solution exists and is unique in Section \ref{sec:dovtv}. Section \ref{sec:bregman} describes the numerical scheme and Section \ref{sec:experiments} presents the numerical evaluation with applications in image denoising, inpainting and deblurring. Section \ref{sec:conclusion} concludes the paper. 

Next we review total variation methods and present current generalizations to color image processing before we introduce our framework.

\section{Further Related Work}
\label{sec:hist_TV}
In addition to TV, closest to our work is the seminal work of Sapiro and Ringach \cite{541429} who first observed that the metric tensor eigendecomposition can be used to describe directional change and magnitude of color images. In this work we extend this reasoning and show that there exists a natural color space representation which leads to a corresponding interpretation of the Sapiro and Ringach approach. Unlike the Beltrami flow \cite{Kimmel2000} and Sapiro and Ringach, we exploit the inverse rate of change of the metric tensor's eigenvalues. This gives us a transformation from the double-opponent space back to the observation space of the image data. We thus obtain an explicit information about the image chromaticity, and by extension, the color edge information. The detection of edges is a well investigated field of study for gray-scale images and methods include, \eg, the canny edge detector \cite{4767851}, gradient filters and the structure tensor \cite{Forstner87,diva2:274026}. These methods work well for monochromatic images (such as gray-scale images) but the extension to multi-dimensional data such as color image data is still an open problem. One of the first extension of the structure tensor to multi-valued images was proposed by Di Zenzo \cite{DiZenzo1986116}, but later it was reported that channel-by-channel denoising is sufficient in the framework of partial differential equations, \eg, \cite{WeickertCoherence1999}. Coupling of the color channels were investigated in \cite{Tschumperle:2003:VIR:1965841.1965926} and decorrelation approaches to denoising have also been considered see, \eg, \cite{diva2:543914}. 

For a gray-scale image $u : \Omega \rightarrow \mathbb{R}$ defined on a domain $\Omega \subset \R^{2}$, the total variation measure is given by
\begin{equation}
  \label{eq:TVc1}
  TV(u) = \sup \left \{ \int_\Omega u \div{\varphi} \; dx \colon \varphi \in \mathcal{C}^1_c(\Omega,\mathbb{R}^2), \|\varphi\|_{\infty} \leq 1 \right \}
\end{equation}
A function $u\in L^1(\Omega)$ belongs to the space of functions of bounded variation $BV(\Omega)$ if
\begin{equation}
  \label{eq:BV} 
  \| u \|_{BV(\Omega)} = \|u\|_{L^{1}(\Omega)} + TV(u) < \infty.
\end{equation}
$TV(u)$ given by \eqref{eq:TVc1} is a support function in the sense of convex analysis. Thus, combining $TV(u)$ with another (or more) convex functionals enables to apply a wide range of convex programming techniques. An early basic example is \cite{Chambolle2004}. Further common strategies include the primal-dual algorithm \cite{chambolle2011} and the Split-Bregman approach \cite{doi:10.1137/080725891}.

Next, we review generalizations of scalar TV to vector-valued and color images.

\subsection{Color and vector-valued TV}
Let ${u} : \Omega \rightarrow \mathbb{R}^d$, $u(x) = \big(u_1(x), ..., u_d(x)\big)^\T$, denote a vector-valued image. Color images are represented by the three color components red, green and blue, \ie, $d = 3$. In this section we review some extensions of total variation to vector-valued and color images categorized in three main tracks: channel-by-channel, spectral approaches and decorrelation approaches. We briefly mention PDE-based models. 

\textbf{Channel-by-channel.}
The straightforward extension of TV to vector-valued color image regularization is to apply \eqref{eq:TVc1} channel-by-channel. However, as this naive approach neglects any channel-by-channel correlation one of the first extensions was to penalize color edges across channels as suggested by Blomgren and Chan \cite{661180}. They raised several important aspects highlighting the fact that the extension to color is a non-trivial task. First, they argued that the vector-valued TV should not penalize intensity edges, as there can be a shift in color but not in intensity. Secondly, they advocate that the corresponding TV-regularizer should be rotationally invariant in the image space, although this is disputed in \cite{6909917}. Blomgren and Chan propose  
\begin{equation}
  \label{eq:TVnm}
  TV_{BC}(u) = \sqrt{ \sum_{i=1}^M TV(u_i)^2 },
\end{equation}
with the TV term under the sum given by \eqref{eq:TVc1}. However, applying this model to the problem of color image denoising has been shown to produce significant color smearing artifacts due to insufficient preservation of color edges. The reason of this effect is that the model fails to comprehend that the red, green and blue color components are in fact highly correlated. Thus, due to lacking any  coupling between the color channels, the model produces suboptimal results w.r.t.~to color consistency \cite{doi:10.1137/110823766}.  

Bresson and Chan \cite{xavier2008} considered a vector-valued extension of the scalar dual TV formulation. Based on work by Chambolle \cite{Chambolle2004} and Fornasier and March \cite{Massimo2007}, Bresson and Chan presented a coherent framework for vectorial total variation with a study of well-posedness. 
While their formulation generalizes Chambolle's dual of Blomgrens TV semi-norm \eqref{eq:TVnm}, results still exhibit color smearing. 

We refer to \cite{Duran2015} for an additional discussion on discrete vectorial total variation models. We remark that this underlines the complex nature of color image processing and researchers continued to propose alternative total variation color filtering models, as we will discuss below.

\textbf{Spectral approaches.}
One of the first vectorial TV (VTV) schemes that explicitly take color information into account, was introduced by Sapiro and Ringach \cite{541429}. In an intensity image an edge is localized by changes in the image intensity. The novelty introduced by Sapiro and Ringach is that they exploited the metric imposed by the first fundamental form on the image domain, which couples the RGB-channel's \emph{derivatives},  to indicate the presence of color edges. 
The resulting functional is given by
\begin{equation}
  \label{eq:Sapiro}
  TV_S({u}) = \int_\Omega \sqrt{\lambda_+ - \lambda_-} \; dx,
\end{equation}
where $\lambda_+ > \lambda_- \geq 0$ are the eigenvalues of the metric tensor. Goldluecke and Cremers \cite{5540194} propose a vectorial total variation method based on the largest singular value (hence on the spectral norm) of the derivative matrix $Du$, 
\begin{equation}
  TV_J({u}) = \int_\Omega \sigma_1(D{u}) \; d{x}. 
\end{equation}    
$TV_J$ is closely related to $TV_S$ with the difference that $TV_J$ sets all singular values (except) the largest to zero. Although $TV_J$ improves the signal-to noise ratio, the visual appearance for the denoised results contains considerable color shimmering, visible in homogeneous regions. Thus, despite a coupling between the color channels, the approach still may produce color artifacts.

\textbf{Decorrelation transforms.} Regularization via decorrelation transforms was suggested in \eg, \cite{Chan2001422}. A more recent approach to incorporating color into a total variation formulation was introduced by Ono and Yamada \cite{6909917}. They propose a discrete norm incorporating a weight $w$ between the intensity and chroma in the decorrelation transform ${O}$ (see also \eqref{eq:o1o2o3},\eqref{eq:do-space2} below)
\begin{align}
  \label{eq:Jvtv}
  J_{VTV}({u}) &= \| D_3Ou \|_1^{w} = w \| D_1o_1 \|_1 + \left \| \begin{pmatrix} D_1 o_2 \\ D_1 o_3 \end{pmatrix} \right \|_1
\end{align}
where $\| x \|_1 = \sqrt{\sum_i x_i^2}$. $D_1$ is the derivative matrix for one channel and $D_3 = \diag (D_1,D_1,D_1)$ the three channel derivative matrix. The constant $w \in (0,1)$ determines the weighting between the intensity and the chromaticity of the color space. A smaller $w$ will penalize the chroma. This formulation, however, does not take into account that the subspace defined as the chroma ($o_2,o_3$) is not decorrelated but actually consists of the components hue and saturation. The framework does not respect the non-uniformity of the opponent space. As a consequence, direct regularizing on the chroma via an Euclidean distance metric violates the non-Euclidean structure of this opponent space. Furthermore, it is easy to construct scenarios where the image saturation changes independently of the hue, thus further motivating why the chroma should be decoupled into hue and saturation \cite{munsell1905color}. 

\textbf{PDE-based models.} There are many PDE-based models for color image filtering. We confine ourselves to referring to \cite{diva2:543914,Tschumperle:2003:VIR:1965841.1965926,WeickertCoherence1999} and to discussing the work of Chambolle \cite{413266} who proposed a partial differential equation (PDE)-based anisotropic diffusion model. This model aims to solve the problem of color constancy (referring to the work by Poggio \cite{Hurlbert1998}). Although Chambolle did not present an energy-based total variation approach, his treatment of the color channels and their smoothing along the image gradient direction is relevant to our subsequent analysis. Chambolle defines a PDE with directional diffusivity ${\xi}$ defined as
\begin{equation}
  \label{eq:Chambolle_xi}
  {\xi} \perp \bigg ( (u_2 - u_3)\nabla u_1 + (u_3 - u_1)\nabla u_2 + (u_1 - u_2)\nabla u_3 \bigg ) = 0,
\end{equation}
reducing smoothing perpendicular to the gradients. The coupling between color channels is explicit: the difference of the intensity level of two color components affects the directional smoothing of the third channel. In practice, $\xi$ is used in the heat equation to inhibit smoothing close to color edges. However, as noted by Sapiro and Ringach \cite{541429}, if two channels are equiluminant and if the third channel has an edge, this edge will remain unaffected by the filter.

Although these ideas were presented more than two decades ago, they did not attract much attention in the image processing community. We will see that our perceptual model is related to \eqref{eq:Chambolle_xi} in that our approach penalizes the pair-wise differences between the image derivatives, not the pair-wise intensity differences. In Section \ref{sec:metric_tensor} we derive a color descriptor which couples the color channels in a natural way derived from the geometry of the double-opponent color transform. We will show that a color channel coupling similar to that of Chambolle, in combination with related ideas to the geometric framework of Sapiro and Ringach \cite{541429}, results in a natural description of the image colorfulness.

\section{Color}
\label{sec:appearance}
Color perception is a well studied area and researchers continue to propose color models. To use the ``correct'' color model is application dependent and a non-trivial problem. In this work we focus on the application of denoising. Next we briefly introduce established principles of color space design and recall some terminology. 

\subsection{Terminology}
\begin{figure}[h]
  \begin{center}
  \includegraphics[width=0.2\textwidth]{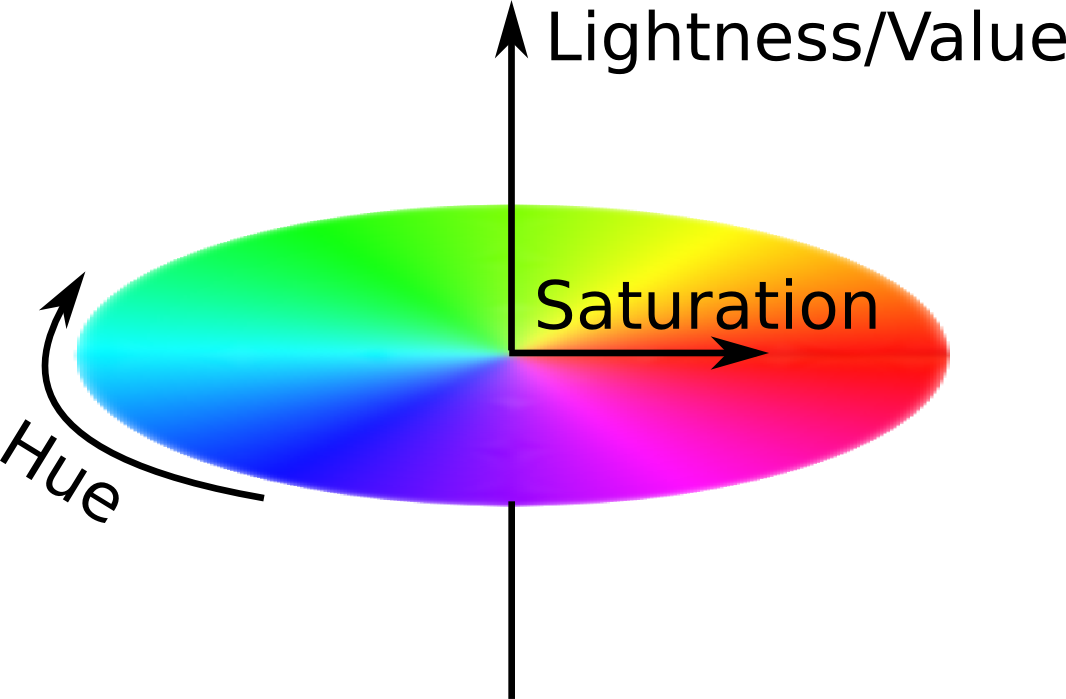}
  \end{center}
  \caption{An isoluminant disc of the doule-opponent color space. Hue is an angular component describing the primary colors red, green, blue, and the opponent colors yellow, magenta and cyan and the corresponding colors transitions. Saturation describes the colorfulness of the corresponding hue and value gives the intensity with which the color is perceived. In the case of scalar valued (gray-scale images) the saturation is null and thus the hue plays no role.}
  \label{fig:colordisc}
\end{figure}

Some of the earliest works on color theory date back to the work by Newton \cite{Newton01011671}. In modern sciences, Albert Munsell is often accredited the notion color dimensions \emph{hue}, \emph{value} and \emph{saturation} \cite{munsell1905color}. Figure \ref{fig:colordisc} illustrates the dimensions where hue is an angular component, value the intensity and saturation a radial component. For consistent use of the color terminology see The Commission Internationale de l'Eclairage (International Commission on Illumination, CIE) \cite{CIE87} and \cite{Sharma:2002:DCI:601332}. For further in-depth information on color image processing and the structure of color we refer to \cite{Sharma:2002:DCI:601332}. When we write intensity, we mean lightness, implying the monochromatic component black and its brightness, or simply, the image gray-scale component. When we write saturation we refer to the magnitude of the color vector orthogonal to the lightness, and hue refers to an angle ranging from 0 to 360 degrees as illustrated in Figure \ref{fig:colordisc}. During the last decades several color systems have been proposed to represent color, each considering different constraint sets. The next section reviews some of the more frequently referred color space representations. 

\subsection{Color space design}
It is important to note when dealing with color theory that color is a subjective experience. In fact, one often thinks of color as wavelength of light. This is a misconception, however, since color is a result of neural processing in the human brain \cite[Ch. 4]{lkp.65448020030101}. This alludes to the difficulty of obtaining a quantitative and accurate description of color. Next, we list four major points that require consideration when determining suitable color space representations. 
 
\begin{itemize}
\item \textbf{Easy to use.} Modeling psychophysical effects of color in scientific applications is a highly non-trivial problem. The study of physical stimulus of different wavelengths has shown that the characterization of color in the human visual system is dominantly done by, what is accepted as the colors \emph{red}, \emph{green}, \emph{blue} and \emph{yellow}. These colors are in fact electromagnetic waves and can be represented as spectral power distributions (SPD) \cite{Sharma:2002:DCI:601332}. This realization of color perception gave rise to one of the earliest color models namely the CIE 1931 XYZ color space which builds on the trichromatic color representation of red (R), green (G) and blue (B). In modern applications these colors constitute the basis of the RGB color space and is widely used to represent image content. In the RGB color space, the relationship between two colors is linear and this simplified representation is also its main drawback. This Euclidean treatment of color relations is not accurate, in fact, as shown in numerous works (see \cite{Sharma:2002:DCI:601332} and references therein), the human visual system is less sensitive to short wavelenghts (\eg, blue), and this therefore suggests that the color spectral distributions should be represented non-uniformly. However, despite obvious oversimplifications in the RGB-representation it may be \emph{sufficiently good} and \emph{easy to use} for the considered application and thus motivates its use. 

\item \textbf{Color reproduction.} Color reproduction is the problem of reproducing color independent of display system. The Munsell system \cite{munsell1905color}, was the first standardization for color metrics widely accepted and later adopted by CIE in 1931. The recommendations set forth by CIE still continue to influence today's color research \cite{Sharma:2002:DCI:601332}. Due to the subjective nature of color perception many color models have been proposed, however, appropriate color space representation seems so far to be application dependent and no consensus has been reached. Due to the emerging technology of Internet and the large variety of imaging display systems and hardware limitations, a standardized RGB color space called sRGB was proposed for consistent image rendering over a wide range of devices \cite{sRGB}. Even up to present time, sRGB is a widely used color space on the world wide web. Also CIELAB is used for color reproduction due to its perceptual uniform structure and is mentioned next. 

\item  \textbf{Perceptual uniformity.} The CIELAB color space was developed for perceptual uniformity. Color and perceptual uniformity, for image processing applications, implies that a perturbation in, \eg, hue, may influence a reference color to a different degree and may induce color artifacts for certain colors, but not for others. Perceptually, this is known to as the just-noticeable-difference and can be visualized via MacAdam ellipses \cite{MACADAM:42}. In reality it has been shown that the CIELAB is \emph{almost} perceptually uniform, see \cite{Sharma:2002:DCI:601332} and references therein. This motivates the use of non-Euclidean metrics, even in supposedly perceptually uniform cases, to determine the distance between colors. Due to the strong emphasis on perceptual uniformity the CIELAB color space is less suitable for rigorous mathematical treatment due to numerous discontinuities it its definition. 

\item  \textbf{Hardware limitations.} Many color spaces were introduced as a consequence of technical and hardware limitations. Such a color space was the HSV (Hue, Saturation, Value) color space developed in the 1970 for applications related to color display systems \cite{Joblove:1978:CSC:800248.807362}. The YCbCr/YPbPr color spaces were proposed for analog and digital television transmission, respectively. Furthermore, the YCbCr/YPbPr signal representation include non-linear mappings and chroma bandlimiting function to enable efficient transmission of the image signal \cite{Poynton2003281}. One can also argue that the previously mentioned sRGB is part of this category. Due to the strong adaptation to efficient engineering requirements, color spaces in this category have not been subject to extensive research in the context of color image processing. 
\end{itemize}

Next we present the double-opponent color space. 

\subsection{Double-opponent color representation}
Recall that the most commonly used color space is the RGB color space. It consists of three components, $r,g$ and $b$ which are the Red, Green and Blue components. These components are uniformly spaced in $[0,255]$ (or $[0,1]$) depending on the chosen quantization. The $r,g,b$ components are highly correlated, and thus image processing algorithms without explicit color adaptation introduce artificial colors and color smearing. To address this problem, a decorrelation transform is usually applied, converting the color space into the three components of lightness,  saturation and hue\footnote{The transformation to these components is \emph{not} equivalent with the HSV color space. Although, the interpretation of the components are similar.}. 

The mapping from the decorrelated double-opponent space in the visual cortex to a physical stimuli is denoted via a function $\psi^{-1} : \mathbb{R}^3 \rightarrow \mathbb{R}^3$ defined by \eqref{eq:psi} below. We denote by, $u : \Omega \rightarrow \mathbb{R}^3$, the physical stimuli containing red, green and blue spectra. Furthermore, denote by the linear mapping 
\begin{equation}\label{eq:o1o2o3}
{O} \colon \R^{3} \to \R^{3},\quad
{u} = (r,g,b)^{\T} \mapsto {O u} = {o}=(o_{1}, o_{2}, o_{3})^{\T},
\end{equation}
the transformation from the RGB color space to the double-opponent space where the matrix $O$ is defined as (see \cite{6751225,Land83,Land86,Ebner:2007:CC:1535382,Lenz2010}),
\begin{equation}
  \label{eq:do-space2}
  {O} =
  \begin{pmatrix}
    1/\sqrt{3} & 0 & 0 \\
    0 & 1/\sqrt{6} & 0 \\
    0 & 0  & 1/\sqrt{2}
  \end{pmatrix}
 \begin{pmatrix}
    1 & 1 & 1 \\
    1 & 1 & -2 \\
    1 & -1  & 0
  \end{pmatrix} .
\end{equation}
We note that this linear mapping has full rank. The matrix ${O}$ actually describes a rotation and scaling of the RGB coordinate system. The opponent component $o_1$ is nothing else than the gray-scale value, $o_2$ is the subtraction of yellow (mixing red and green equals yellow) from blue and the last component $o_3$ is the subtraction of green from red, \ie, $o_2$ and $o_3$ consists of the opponent colors in the RGB color space. 

The non-linear mapping to the hue $(h)$, saturation $(s)$ and lightness $(L)$ representation of the decorrelated double-opponent space is given by
\begin{equation}\label{eq:psi}
  \psi \colon \R^{3} \to \R^{3},\quad
  {o} \mapsto {c} =(L,h,s)^{\T}
\end{equation}
where
\begin{equation}
L = o_{1},\quad h = \arctan\left(\frac{o_{2}}{o_{3}}\right),\quad
s = \left\|\bpm o_{2} \\ o_{3} \epm\right\|,
\end{equation}
Let $\varphi \colon \mathbb{R}^{3} \to \mathbb{R}^{3}$ denote the composition of the linear opponent transform \eqref{eq:do-space2} and the mapping $(o_{1},o_{2},o_{3})^{\top} \mapsto (L,h,s)^{\top}$ just discussed above, then we define 
\begin{equation}
\vphi \colon {u} \to \vphi(u) := \psi({O u}).
\end{equation}
In the next section we compute the metric tensor associated with the mapping $\vphi$ and investigates its encoded information.

\section{Geometry of the double-opponent space}
\label{sec:metric_tensor}

In this section we derive a color representation that allows for an intuitive interpretation of the double-opponent color space as components of the RGB-space. Subsequently we give an exposition on the information that this representation's encodes. 

\subsection{Double-Opponent Metric Tensor}
Strictly speaking, in this work, we regard the RGB-space as a linear Riemannian manifold $\mathcal{M}$ equipped with the metric \eqref{eq:RGB-metric}, which is an inner product on the tangent space $T_{{u}}\mathcal{M}$ that smoothly varies with ${u} \in \mathcal{M}$. Since every tangent space $T_{{u}}\mathcal{M}$ can be identified with $\mathcal{M}$, however, it makes sense to regard the Riemannian metric \eqref{eq:RGB-metric} as inner product defined on the space itself. We refer, e.g., to \cite{Jost2005} for background and further details.

The Euclidean inner product $\langle \cdot,\cdot \rangle$ on the $Lhs$-space induces via $\varphi$ the pullback metric on the RGB-space which is given by
\begin{subequations} \label{eq:RGB-metric}
  \begin{align}
    \langle {u}_{1}, {u}_{2} \rangle_{\bm{u}} 
    &:= \langle D\varphi({u}) {u}_{1}, 
    D\varphi({u}) {u}_{2} \rangle
    = \langle {u}_{1}, G(u) {u}_{2} \rangle, \\ 
    G({u}) &:= \big(D\varphi({u})\big)^{\top} D\varphi({u})
    = \big(g_{ij}({u})\big)_{i,j=1,2,3}.
  \end{align}
\end{subequations}
One identifies that
\begin{subequations}
\begin{align}
u &= (u_{1},u_{2},u_{3})^{\T} = (r,g,b)^{\T}, \\
\alpha &= \alpha(u) := (\alpha_{1},\alpha_{2},\alpha_{3})^{\T}
= (b-g, r-b, g-r)^{\T}, \\
\notag \beta &= \beta(u) := (\beta_{1},\beta_{2},\beta_{3})^{\T} \\
&= (b+g-2 r,b+r-2g, r+g-2b)^{\T}, \\
\notag f^{2} &= f^{2}(u) := \|\alpha\|^{2} \\ &= 2 (b^2-b g-b r+g^2-g r+r^2).
\end{align}
\end{subequations}
In the subsequent analysis we return to the following decomposition
\begin{equation}\label{eq:gamma}
 \gamma = f^2 = {u}^\T{Pu}  
 \qquad 
 {P} = \left(
   \begin{array}{ccc}
     2 & -1 & -1 \\
     -1 & 2 & -1 \\
     -1 & -1 & 2 \\
   \end{array}
 \right) 
 \succeq 0.
\end{equation}
where ${P}$ is a symmetric and positive semi-definite matrix. Furthermore, one easily verifies the relations
\begin{subequations}
\begin{align}
\alpha \perp \beta,\quad
u \perp \alpha,\quad
\alpha \times \beta = f^{2} \eins, \\
\la u, \beta \ra = f^{2},\quad
\|\beta\|^{2} = 3 \|\alpha\|^{2}.
\end{align}
\end{subequations}
The Jacobian $D\varphi$ and the corresponding metric tensor $G$ read
\begin{subequations}
\begin{align}
D\vphi(u) &= 
\dfrac{1}{\sqrt{3}} \bigg(\eins, 
\dfrac{3}{f} \frac{\alpha}{\|\alpha\|},
\dfrac{1}{f} \beta \bigg)^{\T}, \\
G(u) &= \frac{1}{3} \left  (I + \frac{9}{f^{4}} \alpha \alpha^{\T}
+ \frac{1}{f^{2}} \beta \beta^{\T} \right ),
\end{align}
\end{subequations}
where $G$ has non-normalized eigenvectors $\eins, \alpha, \beta$ and corresponding eigenvalues 
\begin{equation}
  \Lambda =  \frac{1}{3}I + \diag \left (0, \frac{3}{f^{2}}, 1 \right ) .
\end{equation}
Saprio and Ringach \cite{541429} also exploited the first fundamental form but in Euclidean space and concluded that the tensor's eigenvalues capture the color edge information. Here we instead use the principal directional change obtained from the eigendecomposition of the  double-opponent metric tensor. And our focus is on $\gamma$ of \eqref{eq:gamma} (which is $f^2$). Similarly to Saprio and Ringach, the interpretation is that a large eigenvalue of the tensor will indicate the presence of image color. Next, we justify and elaborate this statement in the next section while investigating the information encoded in $\gamma$. 

\subsection{Encoded information}
This section investigates the information encoded in the metric tensor, decomposed into an eigensystem as in the previous section. The derived $\gamma$ describes the \emph{colorfulness} of an image, and in the next section we formulate an energy model that explicitly preserve discontinuities in $\gamma$. We identify the following relation
\begin{align}
\notag  \gamma({u}) &= {u}^\T {P} {u} 
  = (b - r)^2 + (r-g)^2 + (g - b)^2 \\
  \label{eq:yQ_F}  &= \| {u}^\T C_1 \|^2, \qquad C_1 =  \begin{pmatrix} 
    1  & -1 & 0 \\
    0  & 1  & -1 \\
    -1 & 0  & 1\\
  \end{pmatrix} .
\end{align}
Note that ${P} = {C_1C_1}^\T$ (cmp. \eqref{eq:gamma}). The coefficients of ${C_1u}$ have previously appeared in an early work by Chambolle \cite{413266}. 

\begin{proposition}\label{prop:p1p2}
  The function, $\gamma$ in \eqref{eq:yQ_F}, has the properties $\mathrm{(P1)} \; \gamma({u} + c \mathbb{1} ) = \gamma({u})$ and $\mathrm{(P2)} \; \gamma(c {u}) = c^2 \gamma({u})$, where $c$ is a constant. 
\end{proposition}
\begin{proof}
  The result follows immediately from \eqref{eq:yQ_F}.
\end{proof}
The above result yields the following interpretation of the $\gamma$-function: \emph{a)} (P1) shows that $\gamma$ is invariant to intensity shifts. \emph{b)} (P2) shows that $\gamma$ has a quadratic dependency on intensity changes. \emph{c)} It follows from \emph{a)} that $\gamma$ depends on color changes. \emph{d)} It follows from \emph{b)} that $\gamma$ depends on color shifts. Under constant intensity, $\gamma$ captures change of color as illustrated by Figure \ref{fig:f} \textbf{(a)}, \textbf{(b)}. In this figure we show equiluminant discs at constant intensity along with the corresponding response of $\gamma$. It is clearly visible that $\gamma$ describes the intrinsic color structure of the color space as there is a stronger response for highly saturated colors. In the lower half of the intensity range we predominantly detect the primary colors red, green and blue. As the intensity increases, $\gamma$ shows primary responses from yellow, cyan and magenta. The intensity axis is located in the center of these discs and, as expected, we do not obtain a value of colorfulness. We give some examples of $\gamma$-responses from natural images in Figure \ref{fig:example_f_images}.

\renewcommand{\imgsize}{0.25\linewidth}
\begin{figure}[t]
  \centering
  \begin{minipage}{0.6\linewidth}
    \centering
    \begin{tabular}{ccc}
      \multirow{1}{*}[8mm]{(\textbf{a})}
      \includegraphics[width=\imgsize]{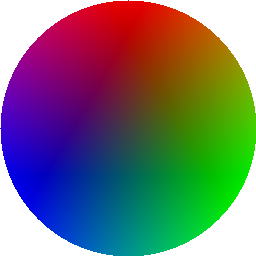} &
      \includegraphics[width=\imgsize]{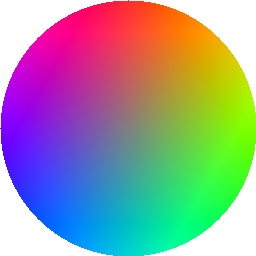} &
      \includegraphics[width=\imgsize]{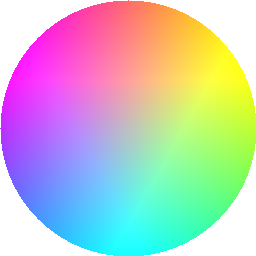} \\
      \multirow{1}{*}[5mm]{(\textbf{b})}
      \includegraphics[width=\imgsize]{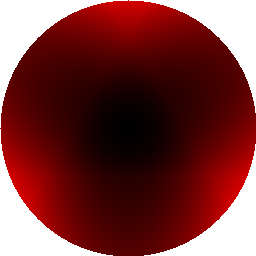} &
      \includegraphics[width=\imgsize]{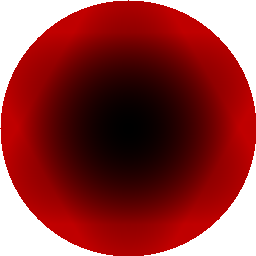} &
      \includegraphics[width=\imgsize]{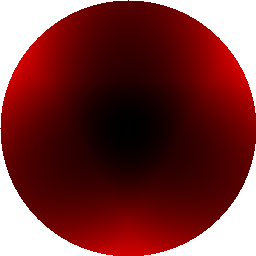} 
    \end{tabular}
  \end{minipage}
  \begin{minipage}{0.32\linewidth}
    \centering
    \hspace{0mm}
    \includegraphics[width=0.75\linewidth]{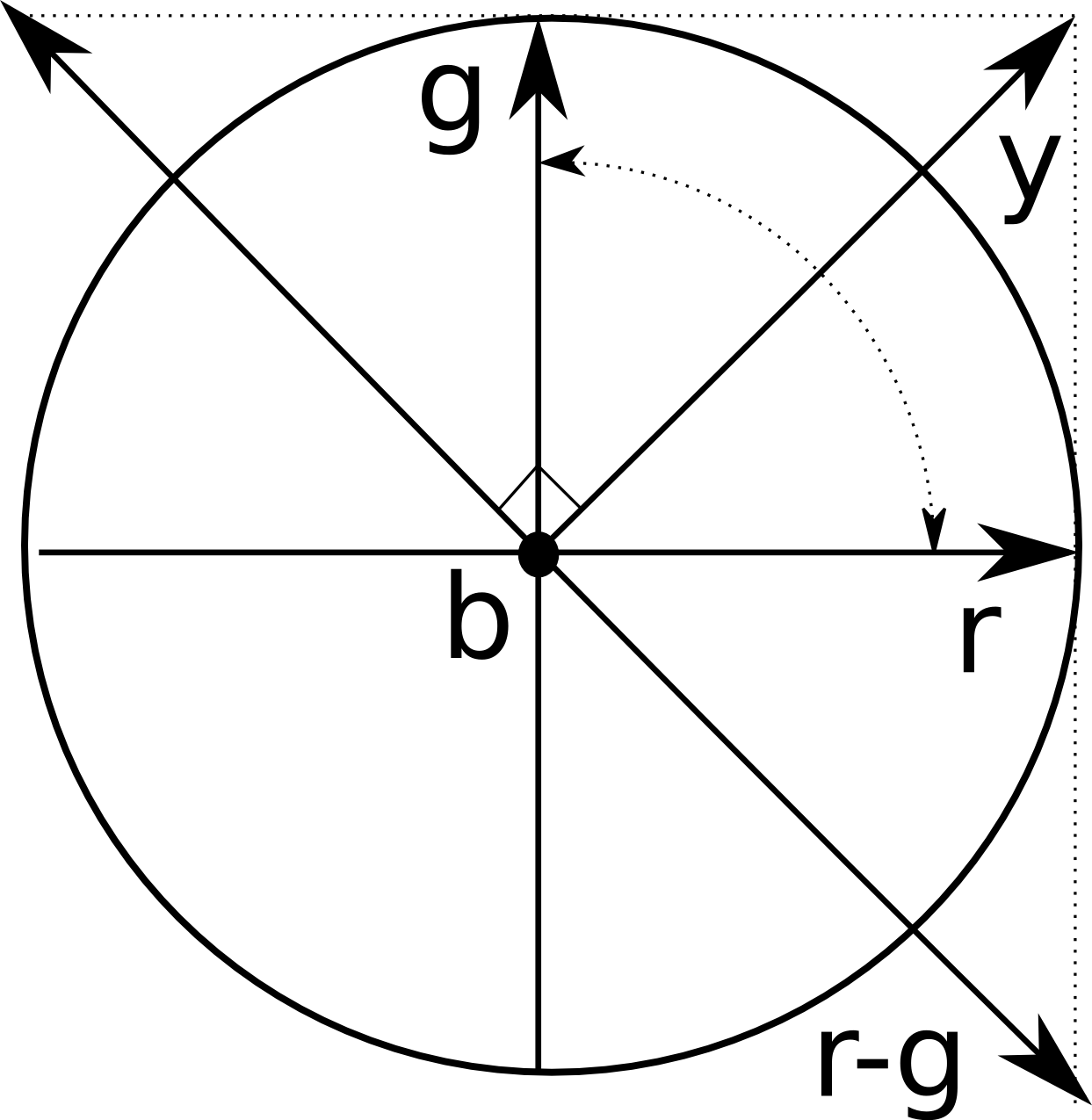} \\
    \hfill \hspace*{0mm} (\textbf{c}) \hfill  \phantom{I} \\
  \end{minipage} \\
  \caption{(\textbf{a}) Color discs with corresponding response of $\gamma$, \eqref{eq:yQ_F}, in (\textbf{b}). The largest magnitude (red color) is obtained at the primary colors (red, green and blue) and the opponent colors (yellow, cyan and magenta). As expected, the response on the intensity axis (center of discs) is 0 (black). (\textbf{c}) Interpretation of the vector $r-g$ as an orthogonal component to yellow. } 
  \label{fig:f}
\end{figure}

The \emph{geometric interpretation} of $\gamma$ is illustrated as an example via the $r-g$-component. The other two color difference terms follow with similar reasoning. We know that the color yellow, $y$, is composed as a sum of red and green, \ie,  $y=r+g$, and written in vector form we have $r-g = (1, -1, 0)^\T$ and $y = r + g = (1, 1, 0)^\T$. We see that yellow is perpendicular to the difference $r-g$, \ie, $y \perp (r-g) = 0$. This is illustrated in Figure \ref{fig:f} \textbf{(c)}. Analogous argument hold for the other terms of $\gamma$, \ie, $b-r$ is orthogonal to magenta, and $g-b$ orthogonal to cyan. In this way $\gamma$ covers the RGB space. Moreover, as $\gamma$ describes the color structure, preserving its edge information prevents color distortion in the filtering process. Based on this analysis, we are now prepared to introduce the general variational formulation.

\section{General Variational Formulation}
\label{sec:energy}
Let  $u = [u_R^\T, u_G^\T, u_B^\T]^\T \in \mathbb{R}^{3N}(\Omega)$ represent a color image on the domain $\Omega$ with stacked color channels. $N$ is the number of pixels in one image channel. Also, let $g \in \mathbb{R}^{3N}$ be the corresponding noisy data. We define the discrete image gradient for \emph{one} channel as $D_1 = \begin{bmatrix} D_x \\  D_y \end{bmatrix} \in \mathbb{R}^{2N\times N}$, $D_x, D_y \in \mathbb{R}^{N\times N}$ and subscript denotes the forward finite difference operator in $x$ and $y$ directions, respectively. Furthermore, we let $i \in \mathbb{N}^+$ s.t.  $I_i \in \mathbb{R}^{i\times i}$ denotes the identity matrix. In this notation, the three channel derivative matrix for a color image is $D^{(1)} = D_1 \otimes I_3 \in \mathbb{R}^{6N\times 3N}$ where $\otimes$ is the Kronecker product and
\begin{equation}
  D^{(1)}u : \mathbb{R}^{3N} \rightarrow \mathbb{R}^{6N}
\end{equation}
is the derivative matrix for the three channels.

\renewcommand{\imgsize}{0.18\linewidth}
\begin{figure}[t]\centering
\begin{tikzpicture}[zoomboxarray, zoomboxes below, zoomboxarray inner gap=0.5cm, zoomboxarray columns=1, zoomboxarray rows=1]
    \node [image node] { \includegraphics[width=\imgsize]{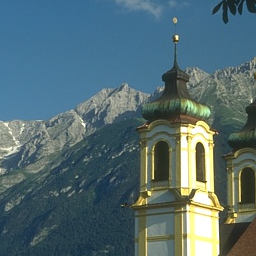} };
    \zoombox[magnification=3,color code=red]{0.7,0.5}
\end{tikzpicture}
\hspace{5mm}
\begin{tikzpicture}[zoomboxarray, zoomboxes below, zoomboxarray inner gap=0.5cm, zoomboxarray columns=1, zoomboxarray rows=1]
    \node [image node] { \includegraphics[width=\imgsize]{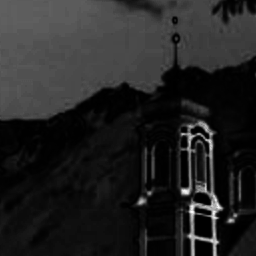} };
    \zoombox[magnification=3,,color code=red]{0.7,0.5}
\end{tikzpicture}
\hspace{5mm}
\begin{tikzpicture}[zoomboxarray, zoomboxes below, zoomboxarray inner gap=0.5cm, zoomboxarray columns=1, zoomboxarray rows=1]
    \node [image node] { \includegraphics[width=\imgsize]{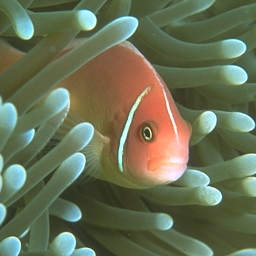} };
    \zoombox[magnification=3,,color code=yellow]{0.7,0.5}
\end{tikzpicture}
\hspace{5mm}
\begin{tikzpicture}[zoomboxarray, zoomboxes below, zoomboxarray inner gap=0.5cm, zoomboxarray columns=1, zoomboxarray rows=1]
    \node [image node] { \includegraphics[width=\imgsize]{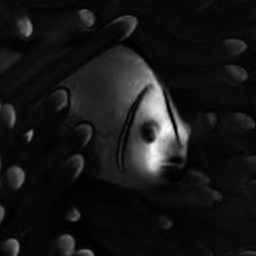} };
    \zoombox[magnification=3,color code=yellow]{0.7,0.5}
\end{tikzpicture}
\caption{Detected color structure in real images extracted by $\gamma$. In these examples, primary colors such as red, green and blue and the opponent color yellow are well characterized.}
\label{fig:example_f_images}
\end{figure}

\subsection{Energy}
Channel-by-channel filtering of the RGB space is prone to introduce color artifacts \cite{661180,xavier2008}. On the other hand, purely decorrelating the color channels without considering the geometry is also sub-optimal, see results in, \eg, \cite{541429,6909917}. We propose a two-component regularizer: one component performs channel-by-channel filtering penalizing \emph{all} intra-channel content and one component which explicitly targets the color information. The color specific prior defines a natural inter-channel coupling from the geometry of the double-opponent space, realized through 
\begin{equation}\label{eq:Cdisc}
  C \in \mathbb{R}^{3N\times 3N}, \quad C = C_1 \otimes I_N 
\end{equation}
where $C_1$ corresponds to \eqref{eq:yQ_F}. 

For $M \in \mathbb{N}$ let $m\in \{1,..,M\}$, then we define $\alpha,\beta \in \mathbb{R}_+^M$ and $p,q \in \mathbb{R}^M(0,2)$, such that $\alpha,\beta,p,q$ are vectors with components indexed by subscript, \eg, $\alpha_m$. We let $s\in (0,2)$ and by $\|x\|_p^p$ we mean
\begin{equation}
  \|x\|^p_p = \sum_{i=1}^{3N} |x_i|^p .
\end{equation}
The energy we introduce and study in this work is defined as the \emph{non-convex} optimization problem 
\begin{equation}\label{eq:mainE}
  \min_{u} \bigg \{ \begin{matrix} E(u) = \frac{1}{s}\|Ku-g\|_s^s \\
  + \sum_{m=1}^M \left ( \frac{\alpha_m}{p_m} \|D^{(m)} u\|_{p_m}^{p_m} +  \frac{\beta_m}{q_m} \| D^{(m)} Cu\|_{q_m}^{q_m} \right ) \end{matrix} \bigg \}, 
\end{equation}
where $D^{(m)}$ is the $m$-order derivative matrix. When $m=1$ we have a \emph{first-order} energy and if $m=2$, we have a second-order energy and so forth. To be explicit, we give the derivative matrix for the first and second order cases 
\begin{subequations}\label{eq:D1D2}
\begin{align}
  D^{(1)} &= \begin{pmatrix}
    D_x \\
    D_y
  \end{pmatrix} \\
  \intertext{and}
  D^{(2)} &= 
  \begin{pmatrix}
    D_{xx} \\
    D_{xy} \\
    D_{yx} \\
    D_{yy}
  \end{pmatrix}
\end{align}
\end{subequations}
where subscript denotes the derivative in $x$ and $y$-directions, respectively. The operator $K \in \mathbb{R}^{3N\times 3N}$ in \eqref{eq:mainE} is the identity matrix in the case of denoising or a blurring matrix in the case of deconvolution. Since the product $Cu$ is now the pair-wise mixture of color channels, it means that $DCu$ is the pair-wise mixture of the channel gradients, \ie, $DCu$ penalizes color opponent gradients in $y,c$ and $m$, and $Du$ penalizes the primary colors $r,g$ and $b$. With $m>1$ we have a natural definition of higher-order total variation. Although $C$ is a constant matrix, it is non-trivial to minimize $E$ due to its inherent non-convexity when either of $p,q,s \in (0,1)$. For this reason, we adopt a half-quadratic formulation presented next. 

\subsection{Half-quadratic formulation}
Alternatives to the HQA \cite{392335} include, \eg, fitting of auxiliary variables \cite{Cohen:1996:AVT:226128.226132} which, however for practical applications, relies on the efficient evaluation of a conjugate function. One may also consider lagged fix-point formulations \cite{doi:10.1137/S0036142997327075}, in this case one would impose regularity to obtain a differentiable energy and prove that there exists a convergent fixed-point algorithm. Instead, the HQA algorithm is particularly suited to optimize $E$ since we obtain a computationally very efficient scheme. Furthermore, identifying that the HQA can be written as an instance of a majorize-minimize scheme we also show that the HQA convergences to a stationary point corresponding to \emph{a} minimum. We make use of the following result to optimize our energy 
\begin{lemma}(HQA $p$-norm \cite{chan2014})\label{lemma:pnorm}
  Let $p \in (0,2)$ and $ t \in (\mathbb{R} \setminus \{0\})$, then  
  \begin{equation}\label{eq:hqa_p}
    \|t\|^p = \min_{v>0} \left \{ vt^2 + \frac{1}{\xi v^\gamma} \right \}
    \end{equation}
    where $\gamma = \frac{p}{2-p}$, $\xi = \frac{2^{2/(2-p)}}{(2-p)p^{p/(2-p)}}$,
    and the minimum is obtained at
    \begin{equation}
      v^* = \frac{p}{2}|t|^{p-2}.
    \end{equation}
\end{lemma}

The energy $E$ in \eqref{eq:mainE} is not differentiable and not convex. To apply the HQA, we make use of the mollified energy $E_\varepsilon$ and set 
$\|x\|_{\eta,\varepsilon}^\eta = \sum_{i=1}^M (|x_i| + \varepsilon)^\eta = \sum_i^M |x_i|_\varepsilon^\eta$ and $0<\eta<2$. Now, the energy $E_\varepsilon(u)$ is differentiable but not convex, therefore the direct optimization problem
\begin{equation}
  u^{k+1} = \min_u E_\varepsilon(u)
\end{equation}
may result in a locally optimal solution. The HQA optimization problem takes the form
\begin{subequations}\label{eq:Eminu}
\begin{align}
\notag & \min_u \Bigg \{ \; E_{\varepsilon}(u) = \sum_{i=1}^{3N}
   \frac{1}{s} \min_{z>0} \left ( z_i |K_iu-g_i|^2_{\varepsilon} + \frac{1}{\xi_s z_i^{\alpha_s}} \right )  \\
   \notag   & + \sum_{m=1}^M \bigg [ \frac{\alpha_m}{p_m} \min_{v_m>0} \left ( v_{i,m} | D_i^{(m)} u |^2_{\varepsilon} + \frac{1}{\xi_{p_m} v_{i,m}^{\alpha_{p_m}}} \right )  \\
& +  \frac{\beta_m}{q_m} \min_{w_m>0} \left ( w_{i,m} | D_i^{(m)} Cu |^2_{\varepsilon} + \frac{1}{\xi_{q_m} w_{i,m}^{\alpha_{q_m}}} \right ) \bigg ] \bigg \}  \\
\label{eq:Eminub}
&  =: \min_{u,v_1>0,...,v_m>0,w_1>0,...,w_m>0,z>0}  \mathcal{L}(u,v,w,z) 
\end{align}
\end{subequations}
now convex separately in $u$ and in $z,v,w$, respectively. We use subscript ``$_i$'' to denote the $i$'th row (or component) of a matrix (or vector). By using Lemma \ref{lemma:pnorm} we formulate an alternating minimization strategy where the update equations for the auxiliary variables are given by
\begin{subequations} \label{eq:wk+1}
\begin{align}
  z^{k+1} &= \frac{s}{2}|Ku^k - g|_{\varepsilon}^{s-2} \\
  ( v_m^{k+1}, w_m^{k+1}) &=  \begin{pmatrix} \dfrac{p_m}{2}|D^{(m)} u^k|_{\varepsilon}^{p_m-2} \\ \dfrac{q_m}{2}|D^{(m)} Qu^k|_{\varepsilon}^{q_m-2} \end{pmatrix}^\T, \; m = 1,...,M
\end{align}
\end{subequations}
and the norm is taken component-wise. The update equation for the \emph{convex} subproblem $u^{k+1}$ is obtained by minimizing 
\begin{align}\label{eq:min_u}
\notag &  u^{k+1} 
= 
\arg \min_u \; \sum_{i=1}^{3N}
  \frac{1}{s} z_i^{k+1}  |K_iu-g |^2_{\varepsilon} \\
   &+ \sum_{m=1}^M \left ( \frac{\alpha_m}{p_m} v_{i,m}^{k+1} | D_i^{(m)} u |^2_{\varepsilon}
   + \frac{\beta_m}{q_m} w_{i,m}^{k+1} | D_i^{(m)} Cu |^2_{\varepsilon} \right ) 
\end{align}

\textbf{Convergence analysis.} The main idea of the following convergence result is to express the HQA as an instance of the majorize-minimize (MMA) algorithm \cite{doi:10.1137/S0036142997327075}. Once we establish the link between the HQA and the MMA we can show convergence of our algorithm. 

Assume there exists a function $\Phi$ such that 
\begin{equation}\label{eq:def_Phi}
   \min_u E_{\varepsilon} (u) = \min_u \Phi(u) 
\end{equation}
and to show that the HQA formulation is of the MMA-type we consider the optimization problem
\begin{equation} \label{eq:def-uk+1}
  u^{k+1} = \arg \min_u \mathcal{F}(u,u^k).
\end{equation}
If there exists a function $\Phi$  such that $\mathcal{F}$ satisfies
\begin{subequations} \label{eq:F-conditions}
\begin{alignat}{4}
\label{eq:cond1MMA}  \mathcal{F}(u,u^k) &\geq \Phi(u),               && \mbox{for} \; \forall u \in \mathbb{R}^n \\
\label{eq:cond2MMA}  \mathcal{F}(u,u^k) &= \Phi(u),                  && \mbox{for} \; u = u^k\\
 \label{eq:cond3MMA}  \nabla_u \mathcal{F}(u,u^k) &= \nabla \Phi(u), \; && \mbox{for} \; u = u^k
\end{alignat}
\end{subequations}
With $\mathcal{L}$ from \eqref{eq:Eminub} and taking into account that $v^{k+1},w^{k+1},z^{k+1}$ depend on $u^{k}$, we define:
\begin{equation}\label{eq:FL}
  \mathcal{F}(u,u^k) := \mathcal{L}(u,v^{k+1},w^{k+1},z^{k+1})
\end{equation}
and have the following result
\begin{proposition}[MMA]
  The optimization problem \eqref{eq:min_u}, stemming from the HQA, is an instance of MMA with $\mathcal{F}$ defined as in \eqref{eq:FL}. 
\end{proposition}

\begin{proof}
  We adopt a the solution strategy introduced in \cite{chan2014}. The first step is to start by substitute \eqref{eq:wk+1} into \eqref{eq:FL} and get expression \eqref{eq:MMA_F}.
 \begin{table*}[b]
    \begin{minipage}{1\linewidth}
         \noindent\makebox[\linewidth]{\rule{\linewidth}{0.4pt}}
      \begin{align}\label{eq:MMA_F}
        \notag \mathcal{F}(u,u^k) & = 
        \sum_{i=1}^{3N} \frac{1}{s} \bigg ( \frac{s}{2}|K_iu^k - g_i|_{\varepsilon}^{s-2} |K_iu-g_i|^2_{\varepsilon}   + \frac{(2-s)}{2} |K_iu^k - g_i|_{\varepsilon}^{s}  \bigg )\\
        \notag & \qquad + \sum_{m=1}^M \bigg [ \frac{\alpha_m}{p_m} \left ( \frac{p_m}{2}|D^{(m)}_iu^k|_{\varepsilon}^{p-2} |D^{(m)}_iu|^2_{\varepsilon} + \frac{(2-p_m)}{2} |D^{(m)}_iu^k|_{\varepsilon}^{p_m} \right ) \\
        & \qquad \qquad + \frac{\beta_m}{q_m} \left ( \frac{q_m}{2}|D^{(m)}_iCu^k|_{\varepsilon}^{q_m-2} |D^{(m)}_iCu|^2_{\varepsilon} + \frac{(2-q_m)}{2} |D^{(m)}_iCu^k|_{\varepsilon}^{q_m} \right ) \bigg ]
      \end{align}
    \end{minipage}
  \end{table*}
 Then we set 
 $\mathcal{F}(u=u^k,u^k)$ which results in
  \begin{align}\label{eq:Phi}
\notag    & \mathcal{F}(u^k,u^k) = \sum_{i=1}^{3N} \frac{1}{2}  |K_iu^k - g_i|_{\varepsilon}^{s}  \\
    &+ \sum_{m=1}^M \Big(\alpha_m  |D_i^{(m)}u^k|_{\varepsilon}^{p_m} 
    + \beta_m   |D_i^{(m)}Cu^k|_{\varepsilon}^{q_m}\Big)  = \Phi(u^k)
  \end{align}
  where $\Phi$ was defined in \eqref{eq:def_Phi} and establishes condition \eqref{eq:cond2MMA}. In order to show condition \eqref{eq:cond1MMA}, \ie, that $\mc{F}$ forms an upper envelope of $\Phi(u)$ we identify that \eqref{eq:MMA_F} can be expressed with Young's inequality. Then under the condition $1/a+1/b = 1$ we have that
  \begin{align}\label{eq:young-ieq}
    \frac{g^a}{a} + \frac{h^b}{b} \geq gh.
  \end{align}
  We set $\xi(u) = |K_iu - g_i|_\varepsilon$ 
  \begin{subequations}
  \begin{align}
    g &= \xi(u^k)^{\frac{(s-2)s}{2}} \xi(u)^s \\
    h &= \xi(u^k)^{\frac{(2-s)s}{2}}          
  \end{align}
  \end{subequations}
  and let $a = \frac{2}{s}$, $b = \frac{2}{2-s}$, which verifies the inequality
  \begin{align}
\notag    & \frac{s}{2}|K_iu^k - g_i|_{\varepsilon}^{s-2} |K_iu-g_i|^2_{\varepsilon} + \frac{(2-s)}{2} |K_iu^k - g_i|_{\varepsilon}^{s}  \\
    & \geq |K_iu - g_i|_{\varepsilon}^{s}, \quad i = 1,...,3N
  \end{align}
  With an analogous reasoning of the remainder of \eqref{eq:MMA_F} components we have that $\mathcal{F}(u,u^k) \geq \Phi(u)$, \ie, condition \eqref{eq:cond1MMA} is fulfilled. Finally, one easily verifies that $\nabla_u \mathcal{F}(u,u^k) = \nabla \Phi(u)$ at $u=u^k$, thus \eqref{eq:cond3MMA} holds. This shows that $\mathcal{F}(u,u^k)$ is a majorizing function of \eqref{eq:Eminu}. 
\end{proof}

\begin{theorem}[Convergence]
  Given a sequence $\{u^k\}$ generated by HQA, then (i) $\Phi(u^k)$, \eqref{eq:Phi}, is monotonically decreasing and convergent and (ii) $\lim \|u^k - u^{k+1}\|_2^2 = 0$ as $k\rightarrow \infty$. 
\end{theorem}
\begin{proof}
The first and second order derivatives of $\mathcal{F}$ are
\begin{align}
\notag  \nabla_u \mathcal{F}(u,u^k)  &=  \sum_{i=1}^{3N} \frac{1}{s} z_i^{k+1} K_i^\T(K_iu-g_i) \\
\notag  & + \sum_{m=1}^M 2\frac{\alpha_m}{p_m}v_{i,m}^{k+1} (D^{(m)})_i^\T D_i^{(m)} u  \\
  & \qquad + 2 \frac{\beta_m}{q_m} w_{i,m}^{k+1} (D^{(m)}_iQ)^\T D^{(m)}_iQ u
\end{align}
and
\begin{align}
\notag  \nabla_u^2 \mathcal{F}(u,u^k)  &=  \frac{1}{s} \sum_{i=1}^{3N} z_i^{k+1} K_i^\T K_i \\
\notag  & + \sum_{m=1}^M 2\frac{\alpha_m}{p_m}v_{i,m}^{k+1} (D^{(m)})^\T D^{(m)} \\
  & \qquad + 2 \frac{\beta_m}{q_m} w_{i,m}^{k+1} (D^{(m)}Q)^\T D^{(m)}Q 
\end{align}
where the latter matrix is positive semidefinite independently of $u$. 
Thus, $\mathcal{F}$ is convex in $u$. 
Moreover, by \eqref{eq:def-uk+1} and \eqref{eq:F-conditions},
\begin{equation} \label{eq:Phi-bounds-F}
  \Phi(u^{k+1}) \leq \mathcal{F}(u^{k+1},u^k) \leq \mathcal{F}(u^k,u^k) = \Phi(u^k).
\end{equation}
From this it is immediate that
\begin{equation}\label{eq:decauPhi}
  \lim_{k\rightarrow \infty} \Phi(u^{k}) - \Phi(u^{k+1}) = 0
\end{equation}
as $\Phi$ is bounded from below by $0$. To show the convergence of $u^k$, we consider the Taylor expansion of $\mathcal{F}(u,u^k)$ at $u^{k+1}$ 
\begin{align}
\notag  \mathcal{F}(u,u^k)  &= \mathcal{F}(u^{k+1},u^k) \\
    &+ \frac{1}{2}(u-u^{k+1})\nabla^2 \mathcal{F}(u,u^k) (u-u^{k+1})
\end{align}
where the first-order term on the right-hand side vanishes due to the optimality condition of \eqref{eq:def-uk+1}. 
Note that higher-order differentials $\nabla_u^{(n)} \mathcal{F}, n>2$ vanish. Minimizing the right-hand side with respect to $u$ and then setting $u=u^{k}$, we obtain
\begin{equation}\label{eq:Fcond2norm}
  \mathcal{F}(u^k,u^k) \geq \mathcal{F}(u^{k+1},u^k) + \frac{\xi}{2}\|u^k-u^{k+1}\|_2^2
\end{equation}
where $\xi$ is positive and is the minimum eigenvalue of $\nabla_u^2\mathcal{F}$. Consequently,  
\begin{equation}
  \mathcal{F}(u^k,u^k) - \mathcal{F}(u^{k+1},u^k) \geq \frac{\xi}{2}\|u^k-u^{k+1}\|_2^2 \geq 0,
\end{equation}
with the left-hand side converging to $0$ due to  \eqref{eq:Phi-bounds-F} and \eqref{eq:decauPhi}.
\end{proof}
Based on the identification with a MMA, we have shown convergence and existence of a solution for the corresponding HQA. Although, this result is significant, we required a mollifier, a constant offset $\varepsilon$ in the denominator. This parameter regularizes the energy and thus introduces smoothness, albeight being small, an returns a smoother solution than desired. In the numerical evaluation we implement the above energy with an iterative Split-Bregman scheme and we obtain stable updates for $\varepsilon$ as small as $10^{-20}$, which we also used in the numerical evaluation. Next, we study the natural selection of first order derivative ($m=1$), quadratic data term ($s=2$) and corresponding vectorial total variation regularization ($p,q=1$) with the same channel coupling matrix as in the HQA. In this case we do not require a mollification and we show that the corresponding minimizer resides in the space of vectorial bounded variation and is unique. 

\subsection{First-Order VTV}
\label{sec:dovtv}
As a special instance of the general variational formulation \eqref{eq:mainE} we study the existence of a solution of a \emph{non-mollified} vectorial total variation term. Consequently we are not required to use the HQA. In this section we will also use a continuous formulation which is more efficient for this purpose, as such, we use the coupling matrix $C_1 \in \mathbb{R}^{3\times 3}$ defined in \eqref{eq:yQ_F}. 

In the following, let $\nabla u = (\nabla u_1, \nabla u_2, \nabla u_3) : \Omega \rightarrow \mathbb{R}^{2\times 3}$ be the \emph{vectorial gradient} of an color image in the generalized sense as discussed in connection with eq.~\eqref{eq:TVc1}, Then we define  ${p} = ({p}_1, {p}_2, {p}_3) \in C_{c}^{1}(\Omega; \mathbb{R}^{2 \times 3})$ with 
$\Div{{p}} = (\div{{p}_1}, \div{{p}_2}, \div{{p}_3})^{\T}$. 
\begin{definition}[Double-opponent VTV] The double opponent regularizer is defined as 
\begin{subequations} \label{eq:def-TVopp}
  \begin{align}
    J_{OPP}(u) &:= \int_\Omega \| \nabla C_1u  \| 
    \\ \label{eq:def-TVopp-b}  
    & := \sup_{\|p\|_{\infty}\leq 1} \left \{ \int_\Omega  \langle C_1u, \Div{p} \rangle  \; dx  \right \}
  \end{align}
\end{subequations}
where $\|{p}\|_{\infty} = \max\{\|{p}_{1}\|,\|{p}_{2}\|,\|{p}_{3}\|\}$.
\end{definition}

Next, we show that the regularizer $J_{OPP}$ is convex, invariant to rotation and intensity shift of the color space.

\begin{theorem}[Invariance and convexity]
  $J_{OPP}$ is rotationally and intensity invariant, 1-homogeneous and convex. 
\end{theorem}
\begin{proof}
Rotational invariance follows from the isotropy of the feasible set of the dual variable ${p}$, that is $\|{p}\|_{\infty}=\|({p}_{1},{p}_{2},{p}_{3})\|_{\infty} \leq 1 \implies \|({R} {p}_{1},{R} {p}_{2},{R} {p}_{3})\|_{\infty} \leq 1$, for any orthogonal matrix ${R}$.
As a consequence of property (P1) and (P2) of Prop.~\ref{prop:p1p2}, $J_{OPP}$ is invariant to intensity shifts, and the relation $J_{OPP}(c {u}) = c J_{OPP}({u})$ is immediate, for any positive constant $c > 0$. Finally, convexity follows from the definition of $J_{OPP}$ as pointwise supremum of affine functions.
\end{proof}
The non-mollified first-order energy of \eqref{eq:mainE}, with a convex dataterm, is defined as 
\begin{equation}\label{eq:minEVTV}
  \min_{{u}} \left \{ \begin{matrix} E(u) = \dfrac{\mu}{2}\|K{u}-{g}\|^2_{L^2(\Omega)} \\[2mm] + \alpha \sum_{i=1}^3 \TV(u_i) + \beta J_{OPP}(u) \end{matrix}  \right \} 
\end{equation}
where $\mu,\alpha,\beta > 0$. 

\vspace{0.25cm}
\noindent
\textbf{Existence of solution.} 
Next we show that the variational approach \eqref{eq:minEVTV} is well posed.

\begin{lemma}[Bounded variation]
  \label{lem:BV}
  Let ${u} \in BV(\Omega;\mathbb{R}^3)$ then ${C_1u} \in BV(\Omega;\mathbb{R}^3)$.  
\end{lemma}
\begin{proof}
Transposing the matrix $C_1$ in the integrand of \eqref{eq:def-TVopp-b} shows that $J_{OPP}(u)$ is the support function of $u$ with respect to the image of the unit ball $\{p \colon \|p\|_{\infty} \leq 1\}$ under the linear mapping $C_1^{\T} \circ \mrm{Div}$. The claim then follows from the assumption $u \in BV(\Omega;\mathbb{R}^3)$.
\end{proof}
As a consequence, the objective function $E({u})$ \eqref{eq:minEVTV} is well defined. We next show that there is a unique color image ${u}$ minimizing $E({u})$.

\begin{theorem}[Uniqueness and existence of solution]\label{thm:existence}
Let ${g} \in L^\infty(\Omega,\mathbb{R}^3)$ and ${u} \in BV(\Omega,\mathbb{R}^3)$. Then there exists a unique minimizer ${u}^{\ast}$ of $E({u})$ given by \eqref{eq:minEVTV}.  
\end{theorem}
\begin{proof} We adapt and sketch a standard proof pattern from \cite{Attouch2014}. Due to ${g} \in L^\infty(\Omega,\mathbb{R}^M)$, we may assume that all admissible ${u}$ are uniformly bounded in the sense that $|u_{i}(x)| \leq \|g_{i}\|_{L^{\infty}(\Omega)},\, i=1,2,3,\, \forall x \in \Omega$. Let $({u}_{n})_{n \in \N}$ be a minimizing sequence with respect to $E({u})$. Then, after passing to a subsequence $({u}_{n_{k}})_{k \in \N}$, there exists a ${u}^{\ast} \in BV(\Omega;\mathbb{R}^3)$ with ${u}_{n_{k}} \to {u}^{\ast}$ strongly in $L^{1}_{loc}(\Omega;\R^{3})$, $\nabla (u_{i})_{n_{k}} \to \nabla u^{\ast}_{i}$ in an appropriate weak sense, and $J_{opp}({u}_{n_{k}}) \to J_{opp}({u}^{\ast})$ in view of Lemma \ref{lem:BV}. It follows from Fatou's lemma and the lower-semicontinuity of $E({u})$ that ${u}^{\ast}$ minimizes $E({u})$, whereas uniqueness of ${u}^{\ast}$ is a consequence of the strict convexity of $E({u})$ due to the data term of \eqref{eq:minEVTV}.
\end{proof}
Next, we derive an efficient numerical scheme which optimizes our proposed energies.

\section{Implementation and Optimization}
\label{sec:bregman}
We briefly comment on convex programming techniques that are relevant to our approach. Then we detail our implementation of a specific technique embedded into the half-quadratic regularization approach.
\subsection{TV and Convex Programming}
There exists numerous methods to minimize the total variation semi-norm. A very popular approach is the primal-dual iteration from Chambolle and Pock \cite{doi:10.1137/S1064827596299767,chambolle2011}. 
Related algorithms include the split bregman method \cite{doi:10.1137/080725891}, augmented Lagrangian methods \cite{Wu:2010:ALM:1958729.1958732} and the alternating direction method of multipliers (ADMM) \cite{Wahlberg549832}. The split-Bregman technique has been shown to be equivalent to ADMM in the case of a linear constraint set \cite{Esser2009}.
Further connections between these methods are discussed in \cite{Wu:2010:ALM:1958729.1958732}.

An evaluation of these algorithms in connection with our approach is beyond the scope of this paper. We  adopted the split-Bregman algorithm \cite{doi:10.1137/080725891} as an established technique and incorporated it as subroutine of our half-quadratic regularization approach, without claiming that this is the best possible choice.

\subsection{Optimization via Split-Bregman and HQA}
With the notation introduced in Section \ref{sec:energy} above, and with the discretized channel matrix $C\in\mathrm{R}^{3N \times 3N}$ (see \eqref{eq:Cdisc}) we write the discretized form of \eqref{eq:mainE} as the optimization problem
\begin{subequations}
\begin{align}
\min_{u,d,e}\; & \frac{\mu}{2}\|Ku-g\|^{2}_2
+ \sum_{m=1}^M \bigg (\alpha \|d_m\|^p_{p} + \beta \|e_m\|^q_{q} \bigg ) \\
&\text{s.t.}\quad
d_m = D^{(m)} u,\quad e_m=D^{(m)} C u.
\end{align}
\end{subequations}
Let $\| v \|_{H}^2 := \langle v, H v \rangle$ be a weighted Euclidean norm and 
$d = (d_1^\T, ..., d_m^\T)^\top$, $e = (e_1^\T, ..., e_m^\T)^\top$. Let 
\begin{equation}
  D = (\vecmat(D^{(1)})^\T, ..., \vecmat(D^{(m)})^\T)^\top
\end{equation}
(cf. \eqref{eq:D1D2}) and set
\begin{align}
  B(u,b,d,e) := \bpm d \\ e \epm -\bpm D \\ D C \epm u - b. 
\end{align}
Applying the Split Bregman approach yields the iteration
\begin{subequations}
\begin{align}
& \notag (u^{k+1},d^{k+1},e^{k+1})
= \min_{u,d,e}\;  
\frac{\mu}{2}\|Ku-g\|^{2}_2  \\
& + \|d\|^p_{p} + \|e\|^q_{q} + \frac{1}{2} \left\|   B(u,b,d,e)   \right\|^{2}_{H},  \\
b^{k+1} &= b^{k} + \bpm D \\ D C \epm u^{k+1} - \bpm d^{k+1} \\ e^{k+1} \epm. \label{eq:b}
\end{align}
\end{subequations}
In the cases of $p=1$ and/or $q=1$ we apply the shrinkage operator to minimize $d,e$, respectively. When $p,q$ are in the non-convex range $(0,1)$, we use the HQA in Lemma \ref{lemma:pnorm} and obtain the point-wise quadratic update step
\begin{subequations}\label{eq:devw}
\begin{align}
\notag   (d^{k+1},e^{k+1}) &= \min_{d,e} v^{k+1} d^2 + w^{k+1}e^2 \\
  & + \frac{1}{2} \left\| B(u^{k+1},b^k,d,e)   \right\|^{2}_{H}, \label{eq:de} \\
  (v^{k+1}, w^{k+1}) & = \Big ( \frac{p}{2}|d^k|^{p-2}_\varepsilon, \frac{q}{2}|e^k|^{q-2}_\varepsilon \Big ), \label{eq:vw}
\end{align}
\end{subequations}
which is strictly convex in both $d$ and $e$. Since \eqref{eq:devw} is defined point-wise, we get the optimality condition, 
\begin{align}
  2\bpm v^{k+1}d  \\ w^{k+1}e \epm + \bpm \alpha & 0 \\ 0 & \beta \epm  \left ( \bpm d \\ e \epm - \bpm D \\ D C \epm u^k - b \right ) = 0 
\end{align}
and the closed-form update-expression
\begin{align} \label{eq:desol}
  \bpm d^{k+1} \\ e^{k+1} \epm  = \bpm (Du^k+b_1) / (1 + 2 v^{k+1}/\alpha) \\  (DCu^k + b_2) / (1 + 2 w^{k+1}/\beta) \epm.
\end{align}

The optimization problem w.r.t. $u$ is solved iteratively by
\begin{align} \label{eq:u-subproblem}
u^{k+1} &= \min_{u}\;\frac{\mu}{2}\|Ku-g\|^{2}_2 
+ \frac{1}{2} \left\| B(u,b^k,d^{k},e^{k}) \right\|^{2}_{H} .
\end{align}
In \eqref{eq:u-subproblem}, we set $b = [b_1^\T, \; b_2^\T]^\T$ for notational convenience and compute
\begin{align}
\notag & \mu K^\top (Ku^{k+1}-g) 
 - \alpha D^\top \Big (d^{k} - D u^{k+1} - b_1^k \Big )  \\
 &- \beta (DC)^\top \Big ( e^{k} - DCu^{k+1} - b_2^k \Big ) 
 = 0
\end{align}
which gives the update step
\begin{align}\label{eq:opt_u}
\notag & \Big (\mu K^\top K  + \alpha D^\top D + \beta (DC)^\top(DC) \Big )u^{k+1}  \\
& = \mu K^\top g + \alpha D^\T(d^k - b^k_1) + \beta (DC)^\T(e^k - b^k_2). 
\end{align}
Our experiments confirm the observation of \cite{doi:10.1137/080725891} that only computing an approximate solution accelerates the overall iterative scheme without compromising convergence. Consequently, we merely apply few conjugate gradient iterative steps to compute $u^{k+1}$. This is computationally cheap since all matrices involved are sparse. 

Finally, we iterate the steps \eqref{eq:opt_u}, \eqref{eq:vw}, \eqref{eq:desol} and \eqref{eq:b} until we satisfy the stopping criterion 
\begin{equation}\label{eq:stoppCriteria}
  \|u^{k}-u^{k+1}\|^2_2/\|u^{k+1}\|^2_2 < 0.9\sqrt{3N\sigma^2}/255^2,
\end{equation}
where $\sigma$ is the noise level standard deviation. This numerical scheme gives stable updates even when $\varepsilon$ is very small. In the experimental evaluation we fixed $\varepsilon = 10^{-20}$. 

\section{Applications}
\label{sec:experiments}

We apply our approach to color image denoising, inpainting and deblurring. All images were normalized to the range $[0,1]$ given in an 8-bit representation. In addition to a qualitative evaluation we include the measures peak signal-to-noise ratio (PSNR), the structural similarity index (SSIM) \cite{1284395} and the CIEDE 2000 which is a measure of color consistency \cite{Sharma2005CIEDE2000}. We consider two denoising scenarios. One synthetic case where the image data consist of piecewise affine regions. The other scenario concerns the denoising of natural images. 

\renewcommand{\imgsize}{0.40\linewidth}
\begin{figure*}[t]
  \centering
  \includegraphics[width=0.38\linewidth]{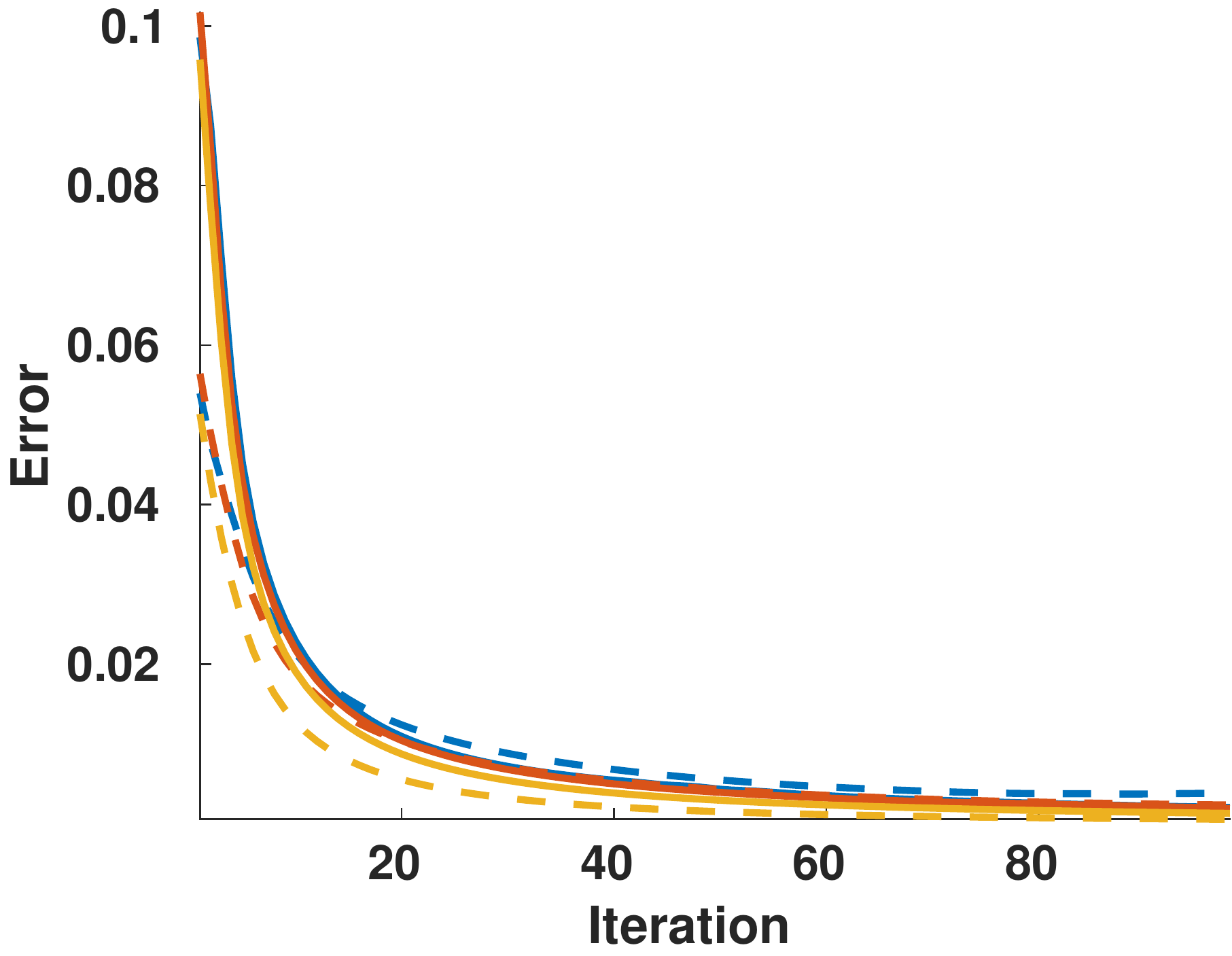} 
  \hspace{10mm}
  \begin{overpic}[scale=0.31,unit=1mm]{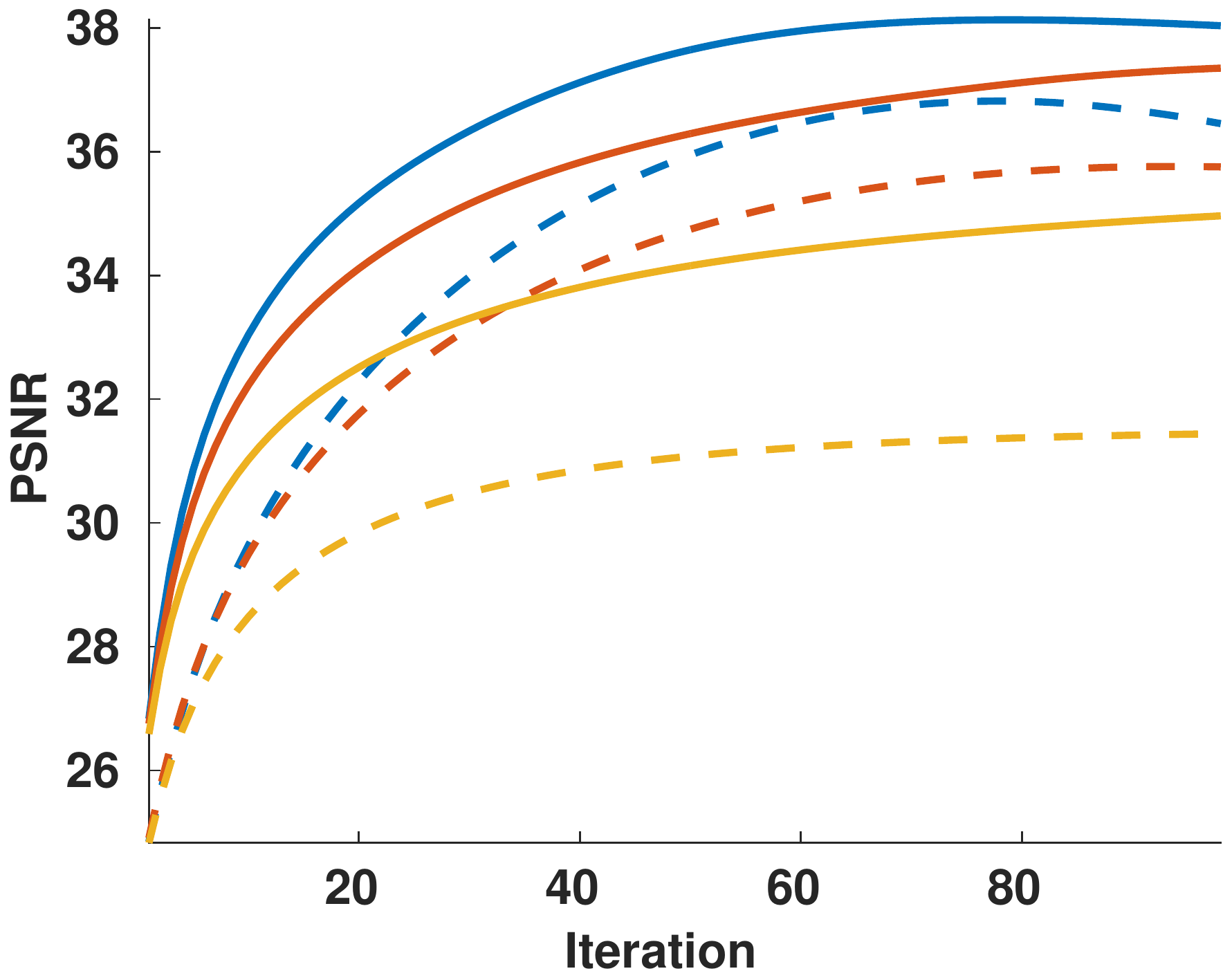}
    \put(-60,45){
      \includegraphics[width=0.15\linewidth]{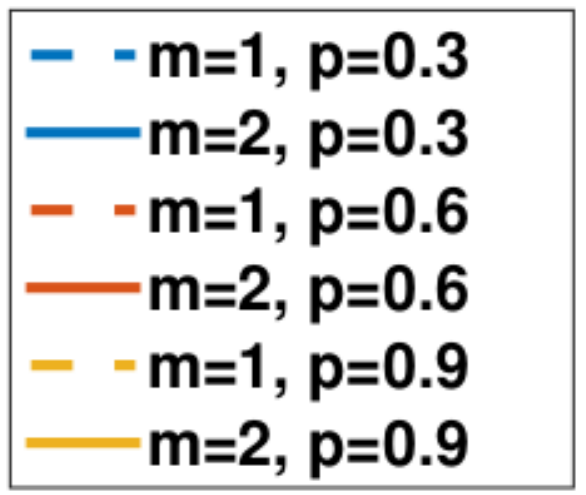}
    }
  \end{overpic}
  \begin{tabularx}{1\linewidth}{XX}
    (a) Normalized difference between two consecutive updates in the iterative update \eqref{eq:opt_u} computed as \eqref{eq:stoppCriteria}. 
    & 
    (b) PSNR value of the current iterate $u^k$. The second order VTV ($m=2$) consistently results in improved PSNR values. 
  \end{tabularx}
  \caption{Empirical convergence (a) of the image data in figure \ref{fig:synthData}. The numerical scheme is implemented as presented in Section \ref{sec:bregman} and the trend of the corresponding PSNR curves are shown in (b). The regularization parameters were set as $\mu=80$, $\alpha_1 = \beta_1 = 2$ for first order VTV. In the second order case we additionally set $\alpha_2 = \beta_2 = 1$. }
  \label{fig:convergencePlots}
\end{figure*} 

\renewcommand{\imgsize}{0.20\linewidth}
\begin{figure*}[t]
  \centering
  \hfill
  \includegraphics[width=\imgsize]{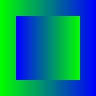} 
  \hfill\hfill
  \includegraphics[width=\imgsize]{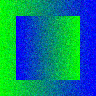}  
  \hfill\hfill
  \includegraphics[width=\imgsize]{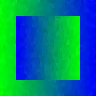} 
  \hfill\hfill
  \includegraphics[width=\imgsize]{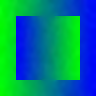}  
  \hfill \phantom{I} \\
  \begin{tabularx}{1.0\linewidth}{XX XX}
    (a) Original
    & 
    (b) Noisy
    &
    (c) First order VTV
    &
    (d) Second order VTV
  \end{tabularx}
  \caption{Synthetic isoluminant test image (a) which consists of affine regions that compose slanting planes in the color space, transitioning from blue to green. (b) shows the noisy data. (c) and (d) depict the first and second order VTV when $p=0.6$. This image illustrates that there is (i) no introduction of artificial colors and (ii) quality of reconstruction is more smooth in the second order VTV compared to the first order VTV due to the higher order constraints on jump-transitions in the image data. Yet the second order VTV shows good edge preservation. The empirical convergence and PSNR values are shown in figure \ref{fig:convergencePlots} (a) and (b). }
  \label{fig:synthData}
\end{figure*}

\subsection{Synthetic image, convergence rate}
\label{sec:syntheticImage}
The synthetic image is isoluminant and therefore presents a particular challenge due to the absence of gray-scale edges. Figure \ref{fig:convergencePlots} shows the empirical convergence rate of our approach for first ($m=1$) and second ($m=2$) order VTV for the noisy image in Figure \ref{fig:synthData}. For each parameter setting the second order VTV shows improved PSNR values which correlates with the visual impression of the final result images in Figure \ref{fig:synthData} (c) and (d), respectively.

\subsection{Denoising of natural images}

The aim of this section is to illustrate differences between commonly benchmarked methods. With this in mind we evaluate the performance of the proposed energy and optimize each method with respect to its parameters and noise levels.  The following methods and parameter ranges are included in the evaluation and we refer to the respective works for further details:
\begin{itemize}
\item Decorrelated VTV  \cite{6909917} (\textbf{DVTV}): Search space for optimal parameter configuration is $\tau \in \{0.95, 1, 1.05\}$, $w \in \{0.3, 0.4, 0.5, 0.6, 0.7\}$. 
\item Primal-dual VTV \cite{xavier2008} (\textbf{PDVTV}): The regularization parameter was optimized for 5 uniformly sampled values in the range $10^{-3}$ to $0.2$.
\item Total generalized variation \cite{doi:10.1137/090769521} (\textbf{TGV}): Applied component-wise and only included for comparison. The regularization parameter was uniformly sampled with 5 values in the range $10^{-3}$ and $0.25$.  
\item Color BM3D \cite{4378954} (\textbf{BM3D}): The standard deviation of the additive Gaussian noise was given as input.  
\end{itemize}

\begin{figure*}[!tbh]
  \centering
  \renewcommand{\imgsize}{0.24\linewidth}
  \newcommand{\xcoord}{1}
  \newcommand{\ycoord}{2.5}
  \overlayZoom{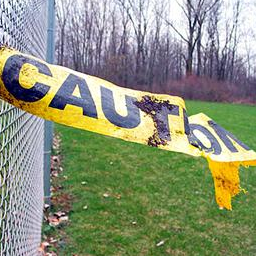}{\xcoord}{\ycoord}
  \overlayZoom{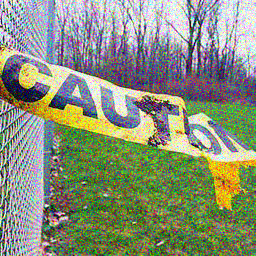}{\xcoord}{\ycoord} \\
  \begin{tabularx}{0.49\linewidth}{XX} 
  \textbf{Original} & \textbf{Noisy} ($60$) \\
  \input{error0_0_840R1_std60.tex}
  \end{tabularx} \\[2mm]
  \overlayZoom{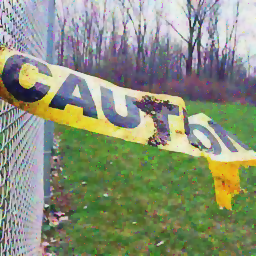}{\xcoord}{\ycoord}
  \overlayZoom{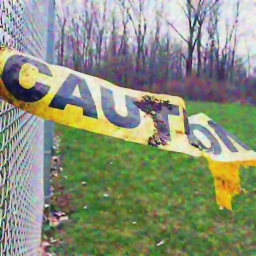}{\xcoord}{\ycoord} 
  \overlayZoom{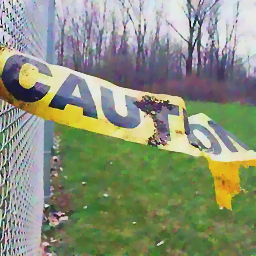}{\xcoord}{\ycoord}
  \overlayZoom{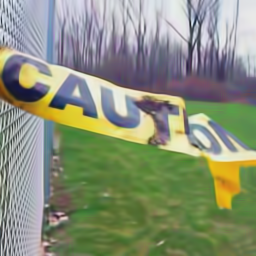}{\xcoord}{\ycoord} \\[3mm]
  \begin{tabularx}{1\linewidth}{XX XX} 
    \textbf{OPP$^1_p$} & \textbf{OPP$^2_p$} & \textbf{OPP$^1_1$} & \textbf{BM3D}  \cite{4378954}\\
    \input{error0_0_840R2_std60.tex}
  \end{tabularx} \\[2mm]
  \overlayZoom{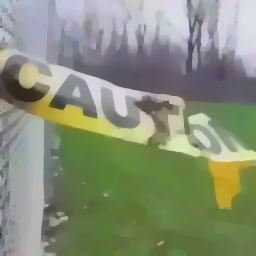}{\xcoord}{\ycoord} 
  \overlayZoom{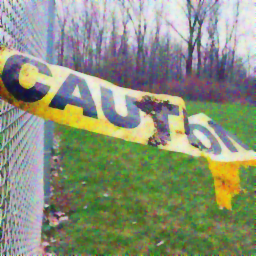}{\xcoord}{\ycoord} 
  \overlayZoom{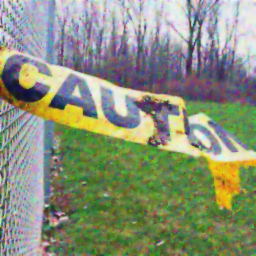}{\xcoord}{\ycoord} 
  \overlayZoom{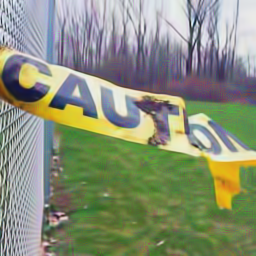}{\xcoord}{\ycoord} \\[1mm]
  \begin{tabularx}{1\linewidth}{XX XX} 
    \textbf{DVTV} \cite{6909917} & \textbf{PDVTV} \cite{xavier2008} & \textbf{TGV} \cite{doi:10.1137/090769521}& \textbf{BM3DS}  \cite{4378954}\\
    \input{error0_0_840R3_std60.tex}
  \end{tabularx} \\
  \vspace{-1mm}
  \caption{Visual comparison of the compared methods and corresponding error values. The result of our OVTV produces the most accurate result, only marginally beaten by BM3D in terms of color accuracy. Yet, the visual quality of OVTV is more clear and does not suffer from desaturated colors as in DVTV. 
  }
  \label{fig:imgdetails}
  \vspace{-3mm}
\end{figure*}

We evaluate our double-opponent (\textbf{OPP}) formulation, see \eqref{eq:mainE} for the case of a first (\textbf{OPP}$^1_p$) and second order energy \textbf{OPP}$^2_p$. We restrict $q=p$ and let $p \in \{ 0.2, 0.6, 0.8, 1 \}$. The following parameter setup was used: 
\begin{itemize}
\item (\textbf{OPP$^1_p$}): First-order scheme optimized with $\mu \in \{ 10,20,30,40,80,100 \}$ and fixed $\alpha_1 = \beta_1 = 2$.
\item (\textbf{OPP$^2_p$}): Second-order scheme optimized with $\mu \in \{ 10,20,30,40,80,100 \}$ and $\alpha_1 = \beta_1 = 2$ and $\alpha_2 = \beta_2 = 1$.
\item (\textbf{OPP$^1_1$}): Optimized parameter space of $\mu$ are 5 uniformly sampled values from $1/255$ to $30/255$, $\alpha = 1$ and $\beta$ was uniformly sampled in from 5 values in the range $1/255$ to $5/255$. 
\end{itemize}

The experimental evaluation use 100 color images from the Weizmann Institute \cite{AlpertGBB07}. The image data were normalized to the range $[0,1]$ using a 8-bit representation.

We are interested in consistent \emph{color} image processing, that is we are specifically tackling the color image filtering problem. Accordingly, we consider the case that the \emph{color} of the image is corrupted by additive Gaussian noise, not the intensity channel. Until now this denoising setup has not been benchmarked and poses an interesting task, \ie, the recovery of \emph{color} information. Rather than corrupting \emph{all} image data with additive noise, we corrupt the components (${o}_2$ and ${o}_3$) with 20, 40, 60 or 80 standard deviations of Gaussian noise and ignore the intensity channel.

After transforming from the (now noisy) opponent representation to the RGB space, one can show that the r,g,b components retain a Gaussian noise distribution as the following calculation illustrates. Let $\sigma^2_i, i = 1,2,3$ denote the standard deviation of zero mean Gaussian noise, then the opponent components, now seen as random variables, are normally distributed, \ie, $o_i \sim \mathcal{N}(0,\sigma_i^2)$. Transformation from the double-opponent space to the RGB color space is done via 
\begin{equation}
  u = O^{-1}o, \qquad 
  O^{-1} = 
  \begin{pmatrix}
    1/\sqrt{3} &  1/\sqrt{6} & 1/\sqrt{2} \\
    1/\sqrt{3} &  1/\sqrt{6} & -1/\sqrt{2} \\
    1/\sqrt{3} &  -2/\sqrt{6} & 0
  \end{pmatrix}.
\end{equation}
Assuming normally distributed opponent components $o \sim \mc{N}(0,\Sigma_{o})$, $\Sigma_{o}=\Diag(0,\sigma^{2}, \sigma^{2})$, we obtain $u \sim \mc{N}(0,\Sigma_{u})$ with 
\begin{equation}\label{eq:sigma_u}
  \Sigma_{u}=O^{-1} \Sigma_{o} O^{-T}
\end{equation}
Excluding the inter-channel correlation between the r,g,b components in \eqref{eq:sigma_u} the noise transformation is given by
\begin{equation}
  u \sim \mathcal{N}(0, \frac{2}{3}\Diag(\sigma^2,\sigma^2,\sigma^2)). 
\end{equation}

Therefore, we also evaluate BM3D with a scaled standard deviation. We denote this approach as 
\begin{itemize}
  \item  Scaled-BM3D (\textbf{BM3DS}): Color BM3D with $\sqrt{2/3}$ of the noise standard deviation.
\end{itemize}
Next we present and discuss the results obtained with OPP and the compared methods.

\begin{figure*}[!tbh]
  \centering
  \renewcommand{\imgsize}{0.20\linewidth}
  \begin{tabularx}{0.83\linewidth}{XX XX} 
    \textbf{Noisy} & \textbf{OPP$^1_p$} & \textbf{OPP$^2_p$} & \textbf{OPP$^1_1$}
  \end{tabularx} \\
  \includegraphics[width=\imgsize]{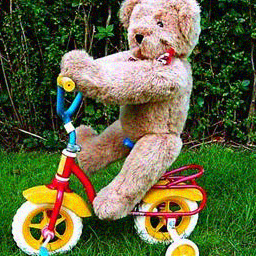} 
  \includegraphics[width=\imgsize]{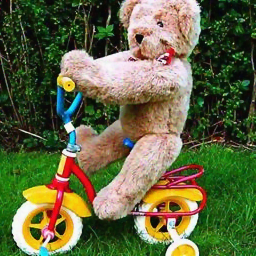} 
  \includegraphics[width=\imgsize]{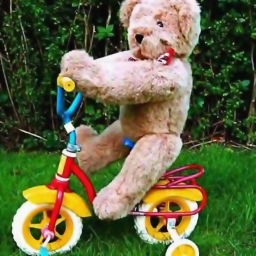} 
  \includegraphics[width=\imgsize]{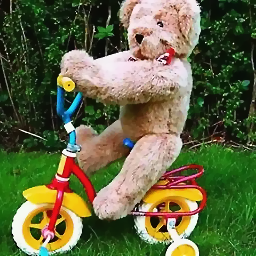} \\[1mm]
  \begin{tabularx}{0.83\linewidth}{XX XX} 
    \input{error0_4_4828_std20.tex}
  \end{tabularx} \\
  \includegraphics[width=\imgsize]{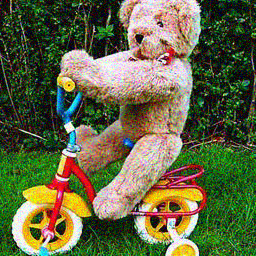} 
  \includegraphics[width=\imgsize]{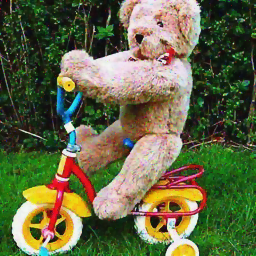} 
  \includegraphics[width=\imgsize]{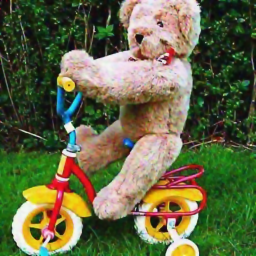} 
  \includegraphics[width=\imgsize]{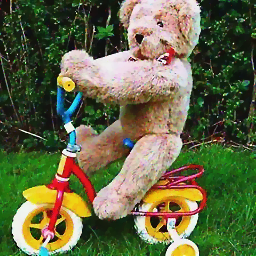} \\[1mm]
  \begin{tabularx}{0.83\linewidth}{XX XX} 
    \input{error0_4_4828_std40.tex}
  \end{tabularx} \\
  \includegraphics[width=\imgsize]{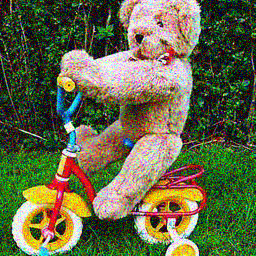} 
  \includegraphics[width=\imgsize]{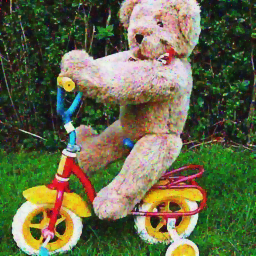} 
  \includegraphics[width=\imgsize]{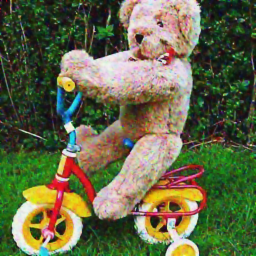} 
  \includegraphics[width=\imgsize]{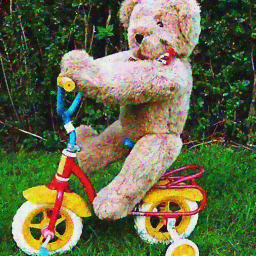} \\[1mm]
  \begin{tabularx}{0.83\linewidth}{XX XX} 
    \input{error0_4_4828_std60.tex}
  \end{tabularx} \\
  \includegraphics[width=\imgsize]{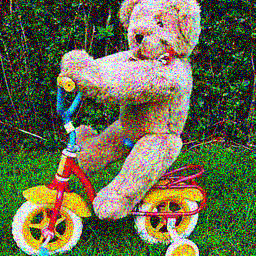} 
  \includegraphics[width=\imgsize]{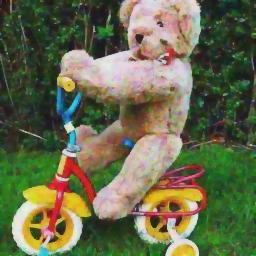} 
  \includegraphics[width=\imgsize]{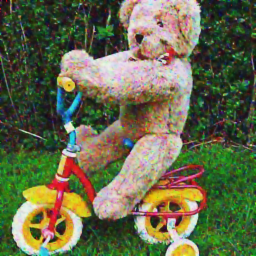} 
  \includegraphics[width=\imgsize]{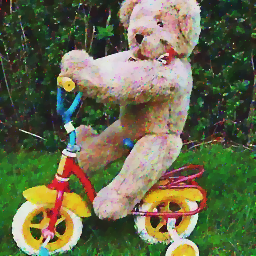} \\ 
  \begin{tabularx}{0.83\linewidth}{XX XX} 
    \input{error0_4_4828_std80.tex}
  \end{tabularx} \\
  \vspace{-1mm}
  \caption{Qualitative comparison between OPP. From top row to bottom row the noise levels the images are corrupted with 20, 40, 60 and 80 standard deviations of noise. For higher noise-levels the second-order OPP$^2_p$ shows improved SSIM, in particular for noise level 80 shows less oversmoothing of the background image structure. Although, it contains more color shimmering reflected in the higher CIEDE value. For lower noise levels, the first-order methods seem preferable. (PSNR/SSIM/CIEDE)}
  \label{fig:varyingNoiseLevel}
\end{figure*}

\begin{figure}[!t]
  \centering
  \renewcommand{\imgsize}{0.3\linewidth}
  \includegraphics[width=\imgsize]{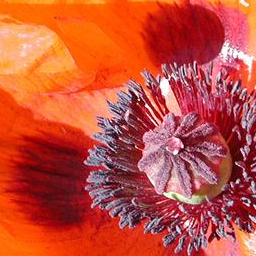}  
  \hspace{1mm}
  \includegraphics[width=\imgsize]{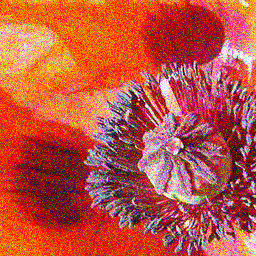}  \\[1mm]
  \begin{tabularx}{0.63\linewidth}{XX} 
    \textbf{Original} & \textbf{Noisy} \\
    &  15.6/0.28/18.71  
  \end{tabularx} \\
  \includegraphics[width=\imgsize]{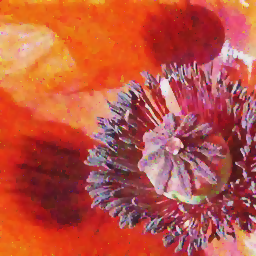} 
  \hspace{1mm}
  \includegraphics[width=\imgsize]{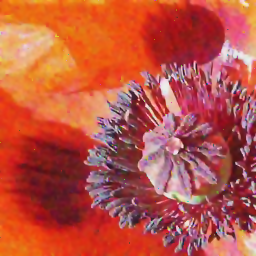} 
  \hspace{1mm}
  \includegraphics[width=\imgsize]{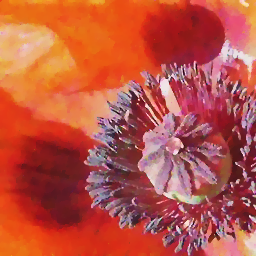} \\[1mm]
  \begin{tabularx}{0.95\linewidth}{XXX} 
    \textbf{OPP$^1_p$}   & \textbf{OPP$^2_p$} & \textbf{OPP$^1_1$} \\
    23.3/0.71/6.02 & 23.7/0.73/5.55  & 24.1/0.74/5.21 
  \end{tabularx} \\
  \vspace{-1mm}
  \caption{Example result images from the evaluation dataset for 60 standard deviations of Gaussian noise. In this image the OPP$^1_1$ performs shows marginally better performance than the non-convex counter-parts. (PSNR/SSIM/CIEDE)
  }
  \label{fig:example1}
\end{figure}

\subsubsection{Results}
Table \ref{table:numResult} shows the average PSNR, SSIM and CIEDE error measures for each method and noise level. Three methods stand out: OPP$^1_p$, OPP$^1_1$ and BM3D. For lower noise-levels the non-convex first-order method OPP$^1_p$ shows highest PSNR and SSIM values. While OPP$^1_1$ performs marginally better than OPP$^1_p$ for noise levels larger than 60 standard deviations of noise, BM3D has the best CIEDE accuracy for all noise levels. Comparing only energy based methods (\ie, not BM3D/BM3DS) it is clear that all OPP-based methods show improved accuracy for PSNR, SSIM and CIEDE in all cases. In all cases OPP$^2_p$ shows worse accuracy than OPP$^1_p$ (except SSIM at noise level 60) due to the extra smoothness constraints imposed by the higher-order derivative. OPP$^2_p$ is more suitable for images with piecewise affine regions, such as illustrated in the synthetic example, Section \ref{sec:syntheticImage}.

The principal difference between OPP$^1_1$ and PDVTV is that the additional color mixing regularization term is included in the former, thus illustrating the success of the mixing term in OPP. With respect to error measures we see an improvement in each case for OPP$^1_1$. Comparing the second order OPP$^2_p$ with the result of TGV the error value differences are smaller, however OPP still shows improved results.

The visual quality, comparing all methods for the (high) noise level 60, is seen in Figure \ref{fig:imgdetails}. All methods suffer from color shimmering in homogeneous regions, although, the shimmering is less obvious for BM3D, which also shows the best color consistency. Although, the result of BM3D is heavily oversmoothed, seen in the detailed panel and the trees in the background. All versions of the proposed OPP  show higher PSNR and SSIM, although in these examples, there is color shimmering. DVTV shows the worst performance and shows strong structural oversmoothing of the trees in the background and there is strong evidence of color smearing. We remark that DVTV is designed for the color transform used in this work, however the method is fundamentally different from ours: DVTV considers the intensity and the color information as orthogonal and independent, and also regularize these separately, however, for the color image data, this is not the case in practice and the method fails as this assumption is violated. 

Figure \ref{fig:varyingNoiseLevel} shows the output for the OPP methods for different noise levels. At the lowest noise-level, the result images appears crisp and do not contain disturbing color artifacts at color edges. Color shimmering artifacts are naturally more pronounced as the noise level increases, due to imprecise parameter settings. Hand-tuning regularization parameters to separate image structure from noise increase clarity and suppress color shimmering, however, further accurate determination of these parameters is subject to further study.

\begin{figure}[!t]
  \centering
  \renewcommand{\imgsize}{0.3\linewidth}
  \includegraphics[width=\imgsize]{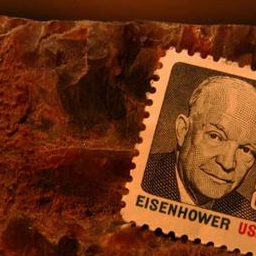} 
  \hspace{1mm}
  \includegraphics[width=\imgsize]{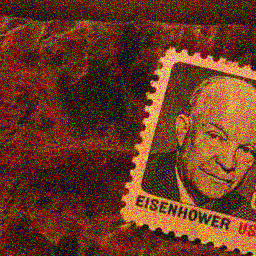}  \\[1mm]
  \begin{tabularx}{0.63\linewidth}{XX} 
    \textbf{Original} & \textbf{Noisy} \\
    & 15.8/0.17/22.44 
  \end{tabularx} \\
  \includegraphics[width=\imgsize]{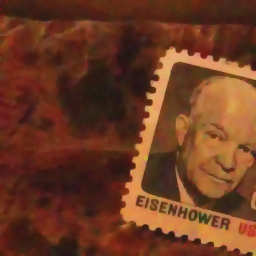} 
  \hspace{1mm}
  \includegraphics[width=\imgsize]{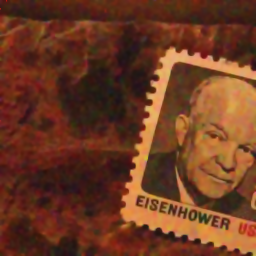} 
  \hspace{1mm}
  \includegraphics[width=\imgsize]{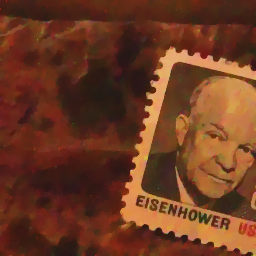} \\[1mm]
  \begin{tabularx}{0.95\linewidth}{XXX} 
    \textbf{OPP$^1_p$} & \textbf{OPP$^2_p$} & \textbf{OPP$^1_1$} \\
    24.9/0.64/5.17 & 25.1/0.65/5.65 & 25.1/0.63/5.27
  \end{tabularx} \\
  \vspace{-1mm}
  \caption{Example result images from the evaluation dataset for 60 standard deviations of noise. OPP$^2_p$ performs well in images with pice-wise affine regions as indicated by the high SSIM value, although with respect to color consistency OPP$^1_1$ shows marginally better result. (PSNR/SSIM/CIEDE) 
  }
  \label{fig:example2}
\end{figure}

To additionally illustrate the recovery of noisy images using OPP, Figures \ref{fig:example1} and \ref{fig:example2} show two challenging images at noise level 60 standard deviations of Gaussian noise. Figure \ref{fig:example1} is an image with vivid colors and many details, visually each methods performs well. OPP$^1_1$ appear most crisp and produces the most smooth background whereas OPP$^{1,2}_p$ both preserve the flower well. Figure \ref{fig:example2} shows similar characteristics as the previous example, although in this case OPP$^2_p$ shows a larger degree of color shimmering than OPP$^1_p$ and OPP$^1_1$. Each of the methods does compensate for the image noise well and preserves facial-features and the stamp's text. 

Table \ref{table:p-results} shows the average error for the used $p$-values in the evaluation of the 100 natural images. There is an indication towards better performance for PSNR and SSIM for $p$ in the range 0.6-0.9 whereas CIEDE suggests smaller $p$-values for the first order scheme. Error values for the second order scheme is in general worse than the first order, this is due to the additional smoothness introduced by the higher-order derivatives. However, as seen in the synthetic result in Section \ref{sec:syntheticImage}, the second-order scheme performs very well when the image data is mostly affine. 

\begin{figure*}[t]
  \centering
  \renewcommand{\imgsize}{0.3\linewidth}
  \includegraphics[width=\imgsize]{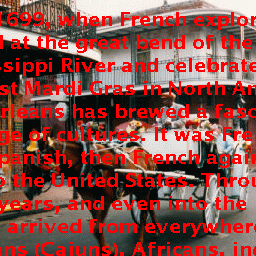} 
  \includegraphics[width=\imgsize]{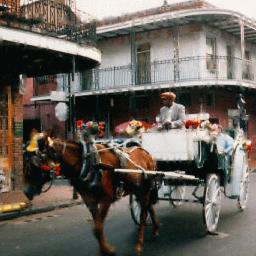} 
  \includegraphics[width=\imgsize]{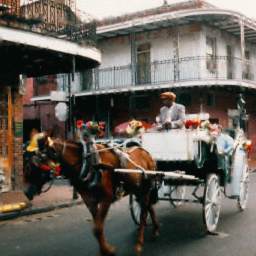}  \\[1mm]
  \begin{tabularx}{0.93\linewidth}{XXX} 
    \textbf{Original} & \textbf{OPP}$^1_{0.8}$ & \textbf{OPP}$^2_{0.8}$  \\
  \end{tabularx} \\
  \vspace{-1mm}
  \caption{The input image contains disturbing text as shown in red color. The restoration task is to remove this text. Final results for the second and first order VTV formulations are seen on the second row for $p=0.8$. Both approaches produces convincing results without introducing disturbing color artifacts or oversmoothing.  
  }
  \label{fig:inpaint}
  \includegraphics[width=\imgsize]{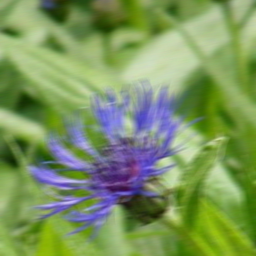}
  \includegraphics[width=\imgsize]{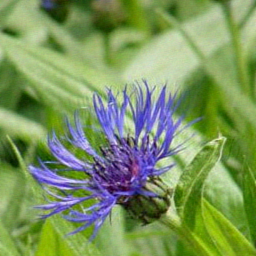}  
  \includegraphics[width=\imgsize]{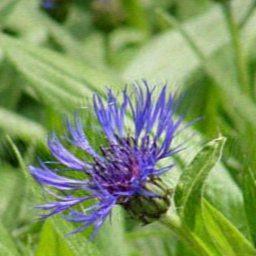}  \\[1mm]
  \begin{tabularx}{0.93\linewidth}{XXX} 
    \textbf{Input}  & \textbf{OPP}$^1_{0.8}$ & \textbf{OPP}$^2_{0.8}$  \hfill {\footnotesize \emph{(PSNR/SSIM/CIEDE)}} \\
  24.20/0.73/3.84 & 32.27/0.93/1.94 & 30.76/0.92/2.01 
  \end{tabularx} \\
  \vspace{-1mm}
  \caption{Example of image sharpening with known motion blur. This example shows that both our first and second order VTV formulations produce high-quality results without introducing disturbing color artifacts, oversmoothing homogeneous regions or over sharpening artifacts.  
  }
  \label{fig:deblurr}
\end{figure*}

\begin{table*}[!t]
  \centering
  \begin{tabularx}{0.95\linewidth}{c ccc ccc}
    & \multicolumn{3}{c}{$\overbracket{\hspace{5.8cm}}^{\displaystyle \mbox{OPP}^1_p}$} & \multicolumn{3}{c}{$\overbracket{\hspace{5.8cm}}^{\displaystyle \mbox{OPP}^2_p}$}  \\[-1mm]
    PSNR/p & 0.3 & 0.6 & 0.9 & 0.3 & 0.6 & 0.9\\
    \toprule
    20 & 32.4 $\pm$ 2.3 & 32.9 $\pm$ 2.0 & 33.0 $\pm$ 1.7 & 29.6 $\pm$ 3.4 & 30.0 $\pm$ 3.2 & 31.1 $\pm$ 2.7 \\ 
40 & 27.5 $\pm$ 2.5 & 28.1 $\pm$ 2.4 & 28.4 $\pm$ 1.9 & 28.0 $\pm$ 2.4 & 28.0 $\pm$ 2.2 & 28.2 $\pm$ 1.9 \\ 
60 & 24.5 $\pm$ 2.3 & 25.3 $\pm$ 2.0 & 25.5 $\pm$ 2.1 & 25.6 $\pm$ 1.9 & 25.6 $\pm$ 1.8 & 25.4 $\pm$ 1.9 \\ 
80 & 22.7 $\pm$ 2.2 & 23.3 $\pm$ 2.1 & 23.4 $\pm$ 1.8 & 23.3 $\pm$ 1.9 & 23.3 $\pm$ 1.8 & 23.2 $\pm$ 1.9 \\ 
 
    SSIM \\
    \toprule
    20 & 0.90 $\pm$ 0.05 & 0.91 $\pm$ 0.05 & 0.91 $\pm$ 0.04 & 0.82 $\pm$ 0.09 & 0.84 $\pm$ 0.07 & 0.88 $\pm$ 0.05 \\ 
40 & 0.79 $\pm$ 0.08 & 0.81 $\pm$ 0.08 & 0.83 $\pm$ 0.07 & 0.80 $\pm$ 0.08 & 0.81 $\pm$ 0.07 & 0.82 $\pm$ 0.06 \\ 
60 & 0.69 $\pm$ 0.11 & 0.74 $\pm$ 0.09 & 0.75 $\pm$ 0.09 & 0.75 $\pm$ 0.08 & 0.75 $\pm$ 0.08 & 0.75 $\pm$ 0.08 \\ 
80 & 0.63 $\pm$ 0.13 & 0.68 $\pm$ 0.11 & 0.69 $\pm$ 0.09 & 0.68 $\pm$ 0.10 & 0.69 $\pm$ 0.09 & 0.69 $\pm$ 0.09 \\ 
 
    CIEDE \\
    \toprule
    20 & 2.99 $\pm$ 2.64 & 2.97 $\pm$ 2.63 & 3.04 $\pm$ 2.64 & 3.42 $\pm$ 2.66 & 3.34 $\pm$ 2.63 & 3.23 $\pm$ 2.62 \\ 
40 & 4.51 $\pm$ 2.54 & 4.40 $\pm$ 2.53 & 4.46 $\pm$ 2.55 & 4.61 $\pm$ 2.51 & 4.69 $\pm$ 2.52 & 4.73 $\pm$ 2.55 \\ 
60 & 5.84 $\pm$ 2.48 & 5.64 $\pm$ 2.42 & 5.68 $\pm$ 2.42 & 5.86 $\pm$ 2.41 & 5.93 $\pm$ 2.43 & 5.95 $\pm$ 2.44 \\ 
80 & 6.90 $\pm$ 2.31 & 6.81 $\pm$ 2.28 & 6.91 $\pm$ 2.28 & 7.09 $\pm$ 2.30 & 7.13 $\pm$ 2.31 & 7.14 $\pm$ 2.32 \\ 

    \bottomrule
  \end{tabularx}
  \vspace{3mm}
  \caption{Error measures and standard deviation for the first and second-order OPP for varying $p$-values. These values were obtained in the optimization of the 100 images in the natural image denoising evaluation. Table \ref{table:numResult} shows the best obtained results in this evaluation scenario.} 
  \label{table:p-results}
  \vspace{0mm}
\end{table*}

\subsection{Image inpainting, image deblurring}
Inpainting refers to the task of restoring \emph{missing} image data, and thus could be viewed as an interpolation problem. Figure \ref{fig:inpaint} illustrates the result for the non-convex OPP for the task of removing the disturbing text in the input image (from \cite{Bertalmio:2000:II:344779.344972}). This is a supervised inpainting task, consequently we define the operator $K$ in our objective function to describe the region of interested to be inpainted, \ie, the text. The second example is the task of restoring image data which has been corrupted by motion blur - a common problem in hand-held imaging devices. In this non-blind deblurring problem we restore the input image, blurred with a motion kernel of 9 pixel shift at and angle of 10 degrees counter-clockwise. The final restoration result, for the first and second order regularization, is seen in Figure \ref{fig:deblurr}. Qualitatively both approaches compensate for the blur and produces similar results in this example image, however the first order scheme shows better error values.

\section{Discussion and Conclusion}
\label{sec:conclusion}

We have shown that the double-opponent theory can significantly improve the performance of vectorial total variation-based methods. Motivated by recent and classical results in color theory, we let the inverse mapping from the opponent-space to the data space serve as a basis of our vectorial formulation. We showed that the inverse mapping encodes image colorfulness. We believe that as the field of perceptual psychophysics continue to evolve we will see further advances in color adaptive algorithms which are inspired by biology. The study presented in this work is a first step towards such models. We formulated a general energy that can model arbitrary higher-order derivatives, where all involved regularization terms can be convex, non-convex or a  mixture. Automatic parameter selection for optimal restoration performance remains an open problem. We discretizised and decomposed our formulation using the iterative split Bregman scheme, and we demonstrated competitive performance compared to standard vectorial total variation methods and state of the art denoising algorithms evaluated using commonly used error measures in the image restoration community. 

\begin{sidewaystable}
  \centering
  \footnotesize
  \begin{tabularx}{0.9\linewidth}{X XXXX XXXX}
    & \textbf{OPP$^1_p$} & \textbf{OPP$^2_p$} & \textbf{OPP$^1_1$} & \textbf{PDVTV} \cite{xavier2008}& \textbf{DVTV} \cite{6909917}   &  \textbf{TGV}  \cite{doi:10.1137/090769521} & \textbf{BM3D}  \cite{4378954} & \textbf{BM3DS}  \cite{4378954}\\ 
    $\sigma=20$  &               &                  &          \\
    \hline \\ [-1.5ex]
    PSNR &\textbf{33.1 $\pm$ 1.85}  & 31.1 $\pm$ 2.72  & 32.0 $\pm$ 2.10  & 29.8 $\pm$ 2.04  & 27.0 $\pm$ 0.99  & 27.3 $\pm$ 2.74  & 32.9 $\pm$ 2.00  & 32.6 $\pm$ 1.19\\ 
SSIM &\textbf{0.92 $\pm$ 0.04}  & 0.88 $\pm$ 0.05  & 0.89 $\pm$ 0.05  & 0.85 $\pm$ 0.05  & 0.79 $\pm$ 0.08  & 0.78 $\pm$ 0.08  & 0.90 $\pm$ 0.05  & 0.87 $\pm$ 0.05\\ 
CIEDE &3.37 $\pm$ 2.70  & 3.31 $\pm$ 2.66  & 2.86 $\pm$ 2.68  & 4.75 $\pm$ 2.91  & 4.57 $\pm$ 2.48  & 5.44 $\pm$ 4.15  & \textbf{2.78 $\pm$ 2.66}  & 3.63 $\pm$ 2.64\\ 

    $\sigma=40$ \\ \hline \\[-1.5ex]
    PSNR &28.4 $\pm$ 2.06  & 28.2 $\pm$ 1.98  & \textbf{28.4 $\pm$ 1.93}  & 25.9 $\pm$ 2.16  & 22.8 $\pm$ 2.17  & 25.4 $\pm$ 2.58  & 28.0 $\pm$ 2.61  & 27.5 $\pm$ 1.75\\ 
SSIM &\textbf{0.83 $\pm$ 0.07}  & 0.82 $\pm$ 0.07  & 0.82 $\pm$ 0.07  & 0.73 $\pm$ 0.08  & 0.60 $\pm$ 0.14  & 0.71 $\pm$ 0.09  & 0.79 $\pm$ 0.09  & 0.74 $\pm$ 0.07\\ 
CIEDE &5.03 $\pm$ 2.69  & 5.33 $\pm$ 2.65  & 5.16 $\pm$ 2.83  & 6.78 $\pm$ 3.06  & 7.12 $\pm$ 2.50  & 6.68 $\pm$ 3.60  & \textbf{4.31 $\pm$ 2.57}  & 5.61 $\pm$ 2.60\\ 

    $\sigma=60$ \\ \hline \\[-1.5ex]
    PSNR &25.5 $\pm$ 2.07  & 25.5 $\pm$ 1.82  & \textbf{25.9 $\pm$ 1.84}  & 23.5 $\pm$ 1.66  & 21.9 $\pm$ 2.21  & 23.6 $\pm$ 2.05  & 25.2 $\pm$ 2.58  & 25.1 $\pm$ 2.07\\ 
SSIM &0.75 $\pm$ 0.08  & \textbf{0.76 $\pm$ 0.07}  & 0.76 $\pm$ 0.09  & 0.60 $\pm$ 0.05  & 0.57 $\pm$ 0.16  & 0.65 $\pm$ 0.10  & 0.71 $\pm$ 0.12  & 0.68 $\pm$ 0.09\\ 
CIEDE &6.46 $\pm$ 2.88  & 6.93 $\pm$ 2.74  & 6.06 $\pm$ 2.57  & 9.22 $\pm$ 2.76  & 7.73 $\pm$ 2.52  & 8.21 $\pm$ 3.17  & \textbf{5.32 $\pm$ 2.46}  & 6.23 $\pm$ 2.46\\ 

    $\sigma=80$ \\ \hline \\[-1.5ex]
    PSNR &23.4 $\pm$ 1.93  & 23.2 $\pm$ 1.82  & \textbf{23.5 $\pm$ 1.82}  & 20.9 $\pm$ 1.01  & 21.0 $\pm$ 2.04  & 21.9 $\pm$ 2.00  & 23.2 $\pm$ 2.37  & 23.2 $\pm$ 2.11\\ 
SSIM &\textbf{0.70 $\pm$ 0.09}  & 0.69 $\pm$ 0.09  & 0.70 $\pm$ 0.09  & 0.45 $\pm$ 0.06  & 0.55 $\pm$ 0.16  & 0.60 $\pm$ 0.11  & 0.66 $\pm$ 0.14  & 0.64 $\pm$ 0.12\\ 
CIEDE &7.60 $\pm$ 2.64  & 8.30 $\pm$ 2.85  & 7.92 $\pm$ 2.78  & 12.75 $\pm$ 2.40  & 8.31 $\pm$ 2.39  & 9.58 $\pm$ 3.46  & \textbf{6.39 $\pm$ 2.34}  & 7.01 $\pm$ 2.29\\ 

    \bottomrule
  \end{tabularx}
  \vspace{3mm}
  \caption{Error measures for the evaluated methods.}
  \label{table:numResult}
  \vspace{-3mm}

\end{sidewaystable}

\addcontentsline{toc}{section}{References}

\bibliographystyle{splncs03}
\bibliography{egbib}

\end{document}